\documentclass[letterpaper,10 pt,conference]{ieeeconf}  

\IEEEoverridecommandlockouts                              
\usepackage{graphicx} 

\usepackage[sort]{cite}
\usepackage{float}
\graphicspath{{./images/}{./images/curves/}}
\usepackage{booktabs}
\usepackage{amsmath} 
\usepackage{amssymb}  
\usepackage[colorlinks=true,linkcolor=red,citecolor=blue,pagebackref=true]{hyperref}

\newlength{\ww}

\newenvironment{noindlist}
 {\begin{list}{\labelitemi}{\leftmargin=0em \itemindent=6pt}}
 {\end{list}}

\newcommand{\A}{\mathbf{A}}
\newcommand{\B}{\mathbf{B}}
\newcommand{\C}{\mathbf{C}}
\newcommand{\D}{\mathbf{D}}

\newcommand{\G}{\mathbf{G}}

\newcommand{\K}{\mathbf{K}}
\newcommand{\bL}{\mathbf{L}}
\newcommand{\I}{\mathbf{I}}
\newcommand{\M}{\mathbf{M}}

\newcommand{\bS}{\mathbf{S}}

\newcommand{\W}{\mathbf{W}}
\newcommand{\X}{\mathbf{X}}
\newcommand{\Y}{\mathbf{Y}}

\newcommand{\bd}{\mathbf{d}}
\newcommand{\be}{\mathbf{e}}

\newcommand{\x}{\mathbf{x}}
\newcommand{\y}{\mathbf{y}}

\newcommand{\one}{\mathbf{1}}
\newcommand{\zero}{\mathbf{0}}

\newcommand{\real}{\mathbb{R}}

\DeclareMathOperator*{\tra}{tr}

\makeatletter
\DeclareRobustCommand\onedot{\futurelet\@let@token\@onedot}
\def\@onedot{\ifx\@let@token.\else.\null\fi\xspace}
\def\eg{\emph{e.g}\onedot} 
\def\ie{\emph{i.e}\onedot}

\def\etal{\emph{et al}\onedot}
\makeatother

\def\ie{\emph{i.e.}}
\def\eg{\emph{e.g.}}
\def\etal{\emph{et al.}}
\def\notation{
	Bold capital letters denote a matrix  $\X$, bold lower-case letters a column vector $\x$.
	$\x_i$ represents the $i^{th}$ column of the matrix $\X$.
	$x_{ij}$ denotes the scalar in the $i^{th}$ row and $j^{th}$ column of the matrix $\X$.
	All non-bold letters represent scalars.
	$\one_{n}\in\Re^{n\times 1}$ is the vector of ones.
	$\I_n \in \real^{n \times n}$ is an identity matrix. 
	$\be_{i} \in \Re^{k}$ is the unit vector with only the $i^{th}$ element is one.
	$\| \X \|_F^2 = \tra(\X^\top \X) = \tra(\X \X^\top)$ designates the Frobenious norm. 
}

\newcommand{\bGamma}{\mathbf{\Gamma}}
\newcommand{\bUpsilon}{\mathbf{\Upsilon}}
\newcommand{\bAlpha}{\boldsymbol{\alpha}}

\title{\LARGE \bf
An Empirical Study of Dimensional Reduction Techniques\\for Facial Action Units Detection
}

\author{
Zhuo Hui and Wen-Sheng Chu \\
Carnegie Mellon University
}

\begin{document}

\maketitle
\thispagestyle{empty}
\pagestyle{empty}

\begin{abstract}


Biologically inspired features, such as Gabor filters, result in very high dimensional measurement.  Does reducing the dimensionality of the feature space afford advantages beyond computational efficiency?  Do some approaches to dimensionality reduction (DR) yield improved action unit detection?  To answer these questions, we compared DR approaches in two relatively large databases of spontaneous facial behavior (45 participants in total with over 2 minutes of FACS-coded video per participant).  Facial features were tracked and aligned using active appearance models (AAM).  SIFT and Gabor features were extracted from local facial regions.  We compared linear (PCA and KPCA), manifold (LPP and LLE), supervised (LDA and KDA) and hybrid approaches (LSDA) to DR with respect to AU detection. For further comparison, a no-DR control condition was included as well.   Linear support vector machine classifiers with independent train and test sets were used for AU detection.  AU detection was quantified using area under the ROC curve and F1. Baseline results for PCA with Gabor features were comparable with previous research. With some notable exceptions, DR improved AU detection relative to no-DR.  Locality embedding approaches proved vulnerable to \emph{out-of-sample} problems.  Gradient-based SIFT lead to better AU detection than the filter-based Gabor features.  For area under the curve, few differences were found between linear and other DR approaches.  For F1, results were mixed.  For both metrics, the pattern of results varied among action units.  These findings suggest that action unit detection may be optimized by using specific DR for specific action units.  PCA and LDA were the most efficient approaches; KDA was the least efficient. 
%
%

\end{abstract}



\section{Introduction}
\label{sec:intro}
\begin{figure}[t]
	\centering
	\includegraphics[width=\linewidth]{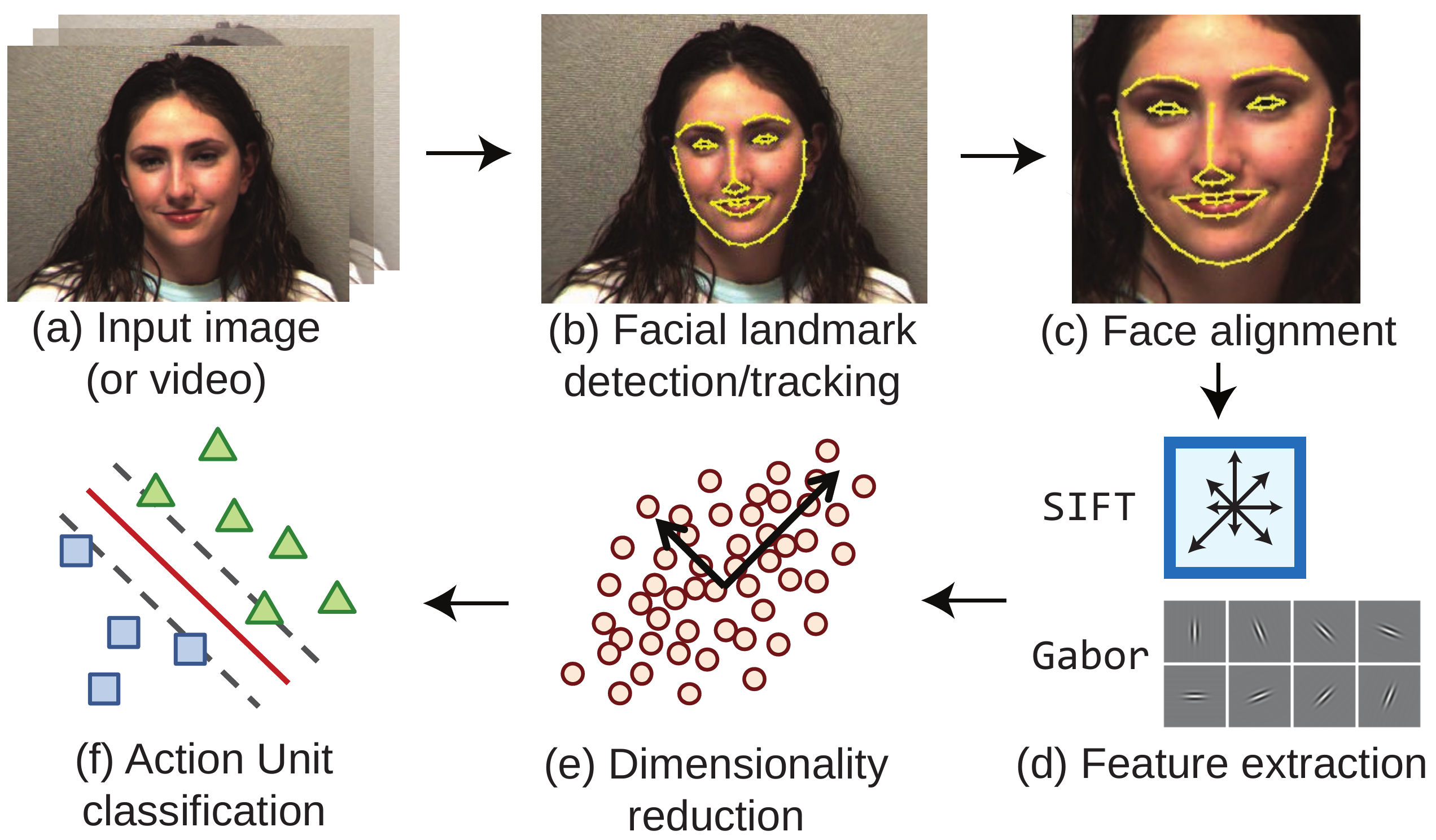}
	\vspace{-18pt}
	\caption{Pipeline for standard facial action unit detection systems.}
	\label{fig:pipeline}
	\vspace{-12pt}
\end{figure}

The face is one of the most powerful channels of nonverbal communication.
The Facial Action Coding System (FACS) \cite{ekman1997face} segments the visible effects of facial muscle activation into Action Units (AU) where each AU is related to one or more facial muscles. 
Because of its descriptive power, FACS has become the state of the art in manual measurement of facial expression and is widely used in studies of spontaneous facial behavior. 
Much effort in facial image analysis seeks to build automatic systems to recognize AUs \cite{pantic2005affective,tian2005facial}.
Fig.~\ref{fig:pipeline} summarizes a standard pipeline for such systems.
Among most systems such as the Facial Expression Recognition and Analysis Challenge (FERA) \cite{valstar2011first}, much attention has been devoted to diverse combinations of feature representations and classification algorithms.

Key issues in automatic facial expression analysis have been addressed in several major reviews.
Fasel and Luettin \cite{fasel2003automatic} reviewed holistic and local strategies for extracting static and temporal facial features and reviewed their efficacy for various classification approaches.
Tian \etal~\cite{tian2005facial} pointed out practical issues regarding an ideal facial expression analysis system, covering the discussions of 2D/3D head pose estimation, shape and appearance feature extraction and frame-/sequence-based classifiers.
Zeng \etal~\cite{zeng2009survey} surveyed multimodal human affective behavior analysis and databases.
More recently, De la Torre and Cohn \cite{de2011facial} provided an overview of state-of-the-art systems, learning procedures (\eg, segment-based classifiers), selection of training samples, and unsupervised discovery of facial events.
We refer interested readers to these references for more details.

A neglected but increasingly salient problem is dimensionality reduction (DR) for use with biologically inspired features.  Initial interest in biologically inspired features resulted from the finding that they outperform shape-only features for many AUs \cite{bartlett2005recognizing,tian2002evaluation}.  More recently, \cite{Chew2012} found that biologically inspired features afford greater robustness to registration error, which is common for spontaneous facial behavior.  The high dimensionality of these descriptors, \eg,~LBP \cite{valstar2011first}, LPQ \cite{jiang2011action}, Gabor \cite{bartlett2006automatic}, and HOG \cite{zhu2011dynamic} entails greater storage and memory demands to perform recognition.  Without improvements in efficiency, the development of practical systems becomes problematic.  Beyond computational efficiency, DR may afford improved accuracy.  


In this paper we emphasize the critical need for DR techniques in AU detection. Our motivation is guided by four observations: 
(1) Human faces are highly structured, especially in comparison with those of other animate and inanimate targets, such as animals or cars.  The structural similarities among human faces are further enhanced  by the constrained poses involved in social interaction, in which people communicate face to face.
(2) Most AU detection systems take video as inputs. Consecutive frames most often carry redundant information.
(3) Uncontrolled environments in real-world settings introduce noise that can produce artifacts or foil detection.
(4) And last, to learn action units in real-wolrd settings requires large numbers of samples at the cost the \emph{curse of dimensionality}.
These observations motivated us to systematically evaluate DR techniques.
By testing the ability of a varied set of DR techniques, we strove to inform and maximize automatic AU detection.

To understand to what extent DR techniques differ in their contribution to AU detection, we implemented a complete AU detection system and systematically varied DR.  We compared exemplars of the full range of DR approaches found in automated facial expression analysis and AU detection.  These are linear, manifold, hybrid, and supervised approaches.   Linear includes Principal Component Analysis (PCA) and Kernel PCA (KPCA). Manifold includes Locality Preserving Projection (LPP) and Locality Linear Embedding (LLE).  Hybrid includes Locally Sensitive Discriminant Analysis (LSDA). Supervised includes Linear Discriminant Analysis (LDA) and Kernel LDA (KDA).  We follow recent advance in formulating popular embedding methods in a least-squares framework \cite{de2011least}.
For baseline, we included a no-DR comparison as well.  Using the unified framework we are able to decompose DR approaches in a systematic way and better analyze their relationships.
This characterization will allow us to bridge the theoretical and practical gaps by comparative experiments.
 

Because our goal is real-world AU detection, we use the realistic datasets, RU-FACS \cite{bartlett2006automatic}, to conduct this comparative study.  The dataset consists of two-person interviews and uses a deception paradigm (RU-FACS) with young adults. The dataset also includes diverse ethnic backgrounds.
The dataset capture varied facial expressions and challenging changes in head pose, illumination, and occlusion. By using large and varied video we can better evaluate DR approaches.
We test the hypothesis that the intrinsic dimension of facial appearance is low (due to the structured appearance of human faces) and constraints in social interaction, and hereby can achieve maximal results by taking into account a low dimensional feature space rather than a whole input space. 
Hopefully, by extracting useful dimensions from complex facial appearance features, one can benefit from lower dimensionality and less noisy data and achieve better efficacy and effectiveness.


\section{Generative Least-Squares Framework for Dimensionality Reduction}
As indicated by recent advance \cite{de2011least}, a unified generative least-squares weighted kernel reduced rank regression (LS-WKRRR) framework can be used to characterize many DR (or component analysis) techniques.
Based on this LS-WKRRR framework, we relate seven representative DR approaches, including linear, nonlinear, hybrid and supervised ones, as particular instances of LS-WKRRR.
We also show that the supervised structrual embedding method, \ie, LSDA \cite{cai2007locality}, that was not included in the original paper \cite{de2011least}, can be also reformulated into an LS-WKRRR.

\begin{figure}[t]
	\footnotesize
	\begin{tabular}{@{}r@{}*{7}{@{}c@{\hspace{4.5pt}}}}
		\toprule
		& {\bf PCA} & {\bf KPCA} & {\bf LPP/LE} & {\bf LLE} & {\bf LDA} & {\bf KDA} & {\bf LSDA} \\
		\midrule
		$\W_r=$ & $\I_d$ & $\I_d$ & $\I_d$ & $\I_d$ & $(\G^\top\G)^{\frac{1}{2}}$ & $(\G^\top\G)^{\frac{1}{2}}$ & $\I_d$\\
		$\bGamma=$ & $\D$ & $\phi(\D)$ & $\phi(\D)$ & $\I_n-\W$ & $\G^\top$ & $\G^\top$ & $\phi(\D)$\\
		$\bUpsilon=$ & $\I_n$ & $\I_n$ & $\I_n$ & $\I_n$ & $\D$ & $\phi(\D)$ & $\I_n$\\
		$\W_c=$ & $\I_n$ & $\I_n$ & $\bS_{\text{LPP}}^{\frac{1}{2}}/\I_n$ & $\I_n$ & $\I_n$ & $\I_n$ & $\bS_{\text{LSDA}}^{\frac{1}{2}}$\\
		\bottomrule
	\end{tabular}
	\vspace{1pt}\\
	* LS-WKRRR formulation: $E_0(\A,\B) = \|\W_r(\bGamma-\B\A^\top\bUpsilon)\W_c\|^2_F$\\
	\vspace{-14pt}
	\caption{Summary of the LS-WKRRR formulation (see more details in text)}
	\label{fig:summary}
	\vspace{-12pt}
\end{figure}

\subsection{The Fundamental LS-WKRRR Formulation}

Given two data sets $\X\in\Re^{x\times n}$ and $\D\in\Re^{d\times n}$ in the input space, and their nonlinear mappings to the high dimensional feature space $\bUpsilon=\phi(\X)\in\Re^{d_x\times n}$ and $\bGamma=\psi(\D)\in\Re^{d_d\times n}$,
the LS-KWRRR \cite{de2011least} unifies various component analysis approaches by the \emph{fundamental formulation} (see notation\footnote{\notation}):
\begin{equation}
	E_0(\A,\B) = \|\W_r(\bGamma-\B\A^\top\bUpsilon)\W_c\|^2_F,
	\label{eq:general}
\end{equation}
where $\W_r\in\Re^{d_x\times d_x}$ is the weighting matrix for the features, and $\W_c\in\Re^{n\times n}$ weights for sample-wise importance.
The regression matrices $\A\in\Re^{d_x\times k}$, $\B\in\Re^{d_d\times k}$ preserves the low-rank correlation between $\bGamma$ and $\bUpsilon$, and spans the column space of $\bGamma$, respectively.
In this study we focus on the DR approaches that can be generalized into this fundamental formulation because:
(1) it yields efficient optimization algorithms to solve the formulation and avoids the common small sample size problem encountered in many DR methods;
(2) it connects different DR approaches in a clean mathmatical formulation, facilitating the understanding the relationship between different types of DR approaches.

\subsection{Linear Methods}

\subsubsection{Principle Component Analysis (PCA)}

PCA is one of the most popular DR techniques \cite{hotelling1933analysis} by keeping the maximal data variance.
PCA has been successfully applied to a large number of face tasks (\eg,~\cite{yang2004two,fasel2003automatic,chu2013selective,zhao2015joint,ding2013facial}), but it may suffer from the drawback that it mainly focuses on retaining large pairwise distance instead of focusing on retaining the small pairwise distances. 
PCA finds an orthogonal subspace $\B\in\Re^{d\times k}$ ($\B^\top\B = \I_k$) that maximizes:
\begin{equation}
	J_{\text{PCA}}(\B) = \tra(\B^\top\C\B),
	\label{eq:PCA}
\end{equation}
where $\C = \frac{1}{n+1}\D(\I_n-\frac{1}{n}\one_n\one^\top_n)\D^\top$ denotes the covariance matrix, and $k$ is the dimension of the subspace (usually determined by preserving its 98\% energy). 
For large amount of high-dimensional data ($d$ and $n$ are large), minimizing the least-squares error function is an efficient procedure (in both space and time) to compute the principal subspace of centered data, \ie, $\D\one_n=\zero$. 
Consider the fundamental Eq.~\eqref{eq:general} where $\bUpsilon = \I_n, \W_r = \I_d, \W_c = \I_n, \bGamma = \D$, PCA can be formulated by minimizing the reconstruction error:
\begin{equation}
	E_{\text{PCA}}(\B,\A) = \|\D-\B\A^\top\|^2_F.
	\label{eq:PCA1}
\end{equation}
The optimal solution can be obtained by the EM-like alternation \cite{roweis1998algorithms} between computing $\A=\D^\top\B(\B^\top\B)^{-1}$ and $\B=\D\A(\A^\top\A)^{-1}$.

\subsubsection{Kernel PCA (KPCA)}
Similar to PCA, KPCA \cite{scholkopf1997kernel} can be derived by lifting the original data samples $\D$ to a feature space, \ie, $\bGamma = \phi(\D)$. 
KPCA has been successfully applied to face recognition \cite{kim2002face} due to its capability to model nonlinear distribution. 
The drawback of KPCA is that the performance may depend on the selected type and parameters of kernel and is computationally expensive. 
The kernelized version of PCA can be written as:
\begin{equation}
	E_{\text{KPCA}}(\B,\A) = \|(\bGamma-\B\A^\top)\|^2_F. 
	\label{eq:KPCA}
\end{equation}
Given an invertible kernel matrix $\K\in\Re^{n\times n}$ and express $\B$ as a linear combination of $\bGamma$, \ie, $\B=\bGamma\bAlpha,\ \bAlpha\in\Re^{n\times k}$, \eqref{eq:KPCA} can be solved alternatively between computing $\bAlpha=\A(\A^\top\A)^{-1}$ and $\A=(\bAlpha^\top\K\bAlpha)^{-1}\bAlpha^\top\K$.

\subsection{Manifold Methods}

\subsubsection{Local Linear Embedding (LLE)}
As one of the representative nonlinear methods, LLE \cite{roweis2000nonlinear} finds an embedding that preserves the local structure of nearby patterns in the high-dimensional space. 
LLE has shown its effectiveness in interpreting nonlinear smooth manifold for facial expressions \cite{chang2003manifold}.
However, LLE is restrictive in modeling manifolds that contains holes.
That is, a new unseen sample is difficult to project onto the embedding learned from the training set, which is the so-called \emph{out-of-sample} problem.
In Sec.~\ref{sec:exp}, we will demonstrate that the out-of-sample problem leads to unsatisfactory results as for AU detection data sampling is usually not dense enough to capture all possible appearance properties such as illumination or head poses.

Firstly, LLE finds a weight matrix, $\W\in\Re^{n\times n}$, $\W\one_n = \one_n$, that captures the neighbor information by minimizing:
\begin{equation}
	J_{\text{LLE$_1$}}(\W) = \sum^n_{i=1}\|\bd_i- \!\!\!\sum_{j\in\aleph(i)}\!\!\! w_{ij}\bd_j\|^2_2
            = \|\D(\I_n-\W)\|_F^2,
	\label{eq:LLE}
\end{equation}
where $\aleph(i)$ denotes the $k$-nearest neighbors of $\bd_i$ and $\W$ contains $k$ or less non-zero values in each column.
Once $\W$ is calculated, LLE finds the embedding $\Y$ that minimizes:
\begin{equation}
	J_{\text{LLE$_2$}}(\Y) = \sum^n_{i=1}\|\y_i-\!\!\!\sum_{j\in\aleph(i)}\!\!\! w_{ij}\y_j\|^2_2
            = \|\Y(\I_n-\W)\|_F^2
	\label{eq:LLE1}
\end{equation}
where $\Y\one_n = 0, \Y\Y^\top =\I_d$.
According to \eqref{eq:LLE1}, LLE can be interpreted as a particular case of KPCA that finds the smallest eigenvectors on the particular kernel matrix $\M=(\I_n-\W)(\I_n-\W)^\top$.
Note that ISOMAP can be also interpreted as KPCA with special kernel matrices \cite{ham2004kernel}, while in this study we show LLE as an exemplar.

\subsubsection{Locality Preserving Projections (LPP)}
LPP \cite{niyogi2004locality}, similar to LLE, finds a linear graph embedding that 
samples originally in close proximity in the input space remain so in the new space. 
Conventional LPP and its improvement has been applied to face recognition \cite{he2003learning,yang2007globally}.
Different from LLE, LPP is defined over the entire ambient space rather than just on the training samples, which can overcome the out-of-sample problems. 
In particular, LPP parameterizes the embedding $\Y\in\Re^{n\times k}$ with a linear transformation of the data, \ie, $\Y = \D^\top\B$, and maximizes:
\begin{equation}
	J_{\text{LPP}}(\B) = \tra({(\B^\top\D\bS\D^\top\B)}^{-1}\B^\top\D\W\D^\top\B),
	\label{eq:LPP}
\end{equation}
where $\B^\top\D\bS\D^\top\B=\I_k$.
Hence, considering \eqref{eq:general} where $\bGamma = \one_n$, $\W_r = \I_d$ and $\W_c = \bS_{\text{LPP}}^{\frac{1}{2}}$, LPP can also be derived as:
\begin{equation}
	E_{\text{LPP}}(\B,\A) = \|(\bGamma-\B\A^\top\D)\bS_{\text{LPP}}^{\frac{1}{2}}\|^2_F,
	\label{eq:LPP1}
\end{equation}
where $\bS_{\text{LPP}}\in\Re^{n\times n}$ is a diagonal matrix of the sum of the rows of $\W$, \ie, $s_{ii}=\sum_j w_{ij}$.
From \eqref{eq:LPP}, LPP can be understood as a reduced rank regression problem from the input space to the feature space with the each sample weighted by $\bS^{\frac{1}{2}}$.
Note that Leplacian Eigenmaps \cite{belkin2001laplacian} also finds a nonlinear embedding to preserves the local structure and can be shown as a particular case of LPP by replacing $\D=\I_n$ \cite{bengio2004learning}.
In this paper we only focus on LPP.


\subsection{Supervised Methods}
\subsubsection{Linear Discriminant Analysis (LDA)}
As a supervised algorithm, LDA \cite{fisher1938statistical} computes a linear transformation ($\A\in\Re^{d \times k}$) of $\D$ that maximizes the Euclidean distance between the means of the classes ($\bS_b$) while minimizing the within-class variance ($\bS_w$). 
LDA has been shown successful to deal with face recognition \cite{deng2005new}, but may suffer from small sample size problem, and the problem will be demonstrated in experiment part. 
LDA can be obtained by maximizing:
\begin{equation}
	J_{\text{LDA}}(\B) = \tra\left((\A^\top\bS_t\A)^{-1}\A^\top\bS_b\A\right),
	\label{eq:LDA}
\end{equation}
where $\bS_b$, $\bS_t$ are the between-class covariance and total covariance, respectively.
Given $\G\in\Re^{n\times d}$ an indicator matrix such that $\sum_j g_{ij}=1$, $g\in\{0,1\}$ indicating whether $\bd_i$ belongs to class $j$, LDA can be formulated as:
\begin{equation}
	E_{\text{LDA}}(\B,\A) = \|{(\G^\top\G)}^{-\frac{1}{2}}(\G^\top - \B\A^\top\D)\|^2_F.
	\label{eq:LDA1}
\end{equation}
The transformation $\A$ can be solved by the trace GEP problem:  $\tra((\A^\top\D\D^\top\A)^{-1}\A^\top\D\G(\G^\top\G)^{-1}\G^\top\D^\top\A)$.
Here LDA can be understood as finding a linear mapping from the data $\D$ to the labels $\G$, where the weighting factor $\G^\top\G$ balances the number of samples between classes.

\subsubsection{Kernel LDA (KDA)}
KDA \cite{mika2002kernel} is a nonlinear extension of LDA, and thus KDA is good at handling the features with small dimension. 
KDA can be simply derived from \eqref{eq:LDA1} by replacing $\D$ with its nonlinear version $\bUpsilon$:
\begin{equation}
	E_{\text{KDA}}(\B,\A) = \|{(\G^\top\G)}^{-\frac{1}{2}}(\G^\top - \B\A^\top\bUpsilon)\|^2_F.
	\label{eq:KDA}
\end{equation}

\subsubsection{Locally Sensitive Discriminant Analysis (LSDA)}
LSDA was not included in the original discussion in \cite{de2011least}.
We show LSDA as another particular case of LS-KWRRR.

LSDA \cite{cai2007locality} can be viewed as a \emph{hybrid} type of manifold and supervised methods.
Different from manifold methods that employ one graph to model the geometrical properties in all data, LSDE uses \emph{two graphs} to model the discriminant structure.
Different from supervised or linear methods that estimate the global data statistics, \ie, mean and covariance, LSDA aims to discover the underlying structure where the data lives on or close to a submanifold of the ambient space.

Given a within-graph $\W_w\in\Re^{n\times n}$ and a between-graph $\W_b\in\Re^{n\times n}$, LSDA maximizes:
\begin{align}
	J_{\text{LSDA}}(\B) &= \tra(\B^\top\D(\alpha\bL_b+(1-\alpha)\W_w)\D^\top\B),
	\label{eq:LSDA1}
\end{align}
where $\B^\top\D\bS_w\D^\top\B=\I_n$, $\bL_b=\D_b-\W_b$ is the Laplacian matrix of the between-graph, $\bS_{\{w,b\},ii}=\sum_j\W_{\{w,b\},ij}$ are the diagonal matrices, and $0\le\alpha\le 1$ is a suitable constant.
Observe that problem \eqref{eq:LSDA1} is similar to the form in \eqref{eq:LPP}.
We can hence interpret LSDA as the same formulation as LPP: 
\begin{equation}
	E_{\text{LSDA}}(\B,\A) = \|(\bGamma-\B\A^\top\D)\bS_{\text{LSDA}}^{\frac{1}{2}}\|^2_F,
	\label{eq:LSDA2}
\end{equation}
where $\bS_{\text{LSDA}}=\alpha\bL_b+(1-\alpha)\W_w$ is a weighting matrix that emphasizes the importance whether the instances belong to the same class.

\section {Experiments}
\label{sec:exp}
This section describes the experiments on RU-FACS \cite{bartlett2006automatic} and Spectrum \cite{cohn2009detecting} databases, which represent a more realistic AU detection scenario in terms of various races, ages, head movements, spontaneous expressions, and partial occlusions.
Our goal is to demonstrate in such diverse data which DR methods are beneficial for facial AU detection.

\subsection {Datasets}

Two databases were used to investigate the influence of differences in dimensionality reduction: RU-FACS \cite{bartlett2006automatic} and Spectrum \cite{APA_1994}.  As explained in the next two sections, the observational scenario for both databases was an interview.  The databases differ in age of participants (older in Spectrum), head pose (near-frontal for RU-FACS and about 15 degrees from frontal for Spectrum), type of stress (deception evasion in RU-FACS and depression in Spectrum), inter-observer reliability (unknown in RU-FACS; good in Spectrum), and reliability of coding.  For RU-FACS, only `B' or higher intensity was coded; whereas all levels of intensity were coded for Spectrum.  These differences could contribute to differences in findings between datasets. 

{\bf RU-FACS:} 
Consists of video-recorded interviews of 100 young adults of varying ethnicity. Interviews were approximately two minutes in duration. Head pose was frontal with small to moderate out-of-plane rotation.  We had access to 34 of the interviews, of which video from five subjects could not be processed for technical reasons (\eg, noisy video). Thus, the experiments reported here were conducted with data from 29 participants. Metadata included manual FACS codes for AU onsets, peaks, and offsets. AU were coded if intensity was greater than `A' (\ie, trace, or lowest intensity on a 5-point scale).  Inter-observer agreement for FACS coding in RU-FACS has not been reported.  Because some AU occurred too infrequently, we selected the nine AUs that most occurred.

{\bf Spectrum:}
Participants were 34 adults (67.6\% female, 88.2\% white, mean age 41.6 years) in the Spectrum database \cite{cohn2009detecting} with a current diagnosis of major depressive disorder \cite{APA_1994} as determined using a structured clinical interview \cite{first1995structured}. They were video-recorded during on one or more occasions at 7-week interviews during a semi-structured interview to assess depression severity (Hamilton Rating Scale for Depression, HRSD \cite{hamilton1960rating}. The interviews were recorded using four hardware-synchronized analogue cameras. Video from a camera roughly 15 degrees to the participant's right was digitized into 640$\times$480 pixel arrays at a frame rate of 29.97 frames per second.

Participant facial behavior was manually FACS coded from video by certified and experienced coders. AU onset, apex, and offset were coded for 17 commonly occurring AU. AU for all levels of intensity were coded (\ie, including `A' or trace level intensity).  Overall inter-observer agreement for AU occurrence, quantified by Cohen's Kappa was 0.75, which is considered a good reliability. The current study analyzed nine of these AU (Table I) that are conceptually related to affect and occurred frequently in the database (i.e., more than 5\% of the time).

\subsection {Experimental Setup}

{\bf Tracking and alignment:}
Parameterized appearance models have been proven useful for facial feature alignment.
In our study, we exploited the Active Appearance Models (AAMs) \cite{matthews2004active} that have been proven an excellent tool for aligning facial features with respect to a shape and appearance model.
In particular, the AAM composes of 66 facial landmarks that deform to fit perturbations in facial features (see Fig.~\ref{fig:pipeline}(b)).
After tracking facial features using AAM, the face is registered to an average face, as shown in Fig.~\ref{fig:pipeline}(c). 

\def\ft{We have done preliminary experiments on deciding the best scale for SIFT. Among 12$\times$12, 24$\times$24 and 36$\times$36 patches, we empirically chose the smallest scale since it gives the best overall accuracy.}

{\bf Appearance features:}
Following current success, we explored the use of the SIFT descriptors \cite{zhu2011dynamic,chu2013selective,zhao2015joint} and Gabor filter responses~\cite{lucey2009automatically,tian2002evaluation} as appearance features.
Since AUs happen only on subregions of the face, we extracted facial features only according to a subset of the 66 landmarks, \ie, 9 points for upper face and 7 points for lower face.
For all AUs, SIFT descriptors are built using a square of 12$\times$12 pixels.
Gabor filter responses are computed as the output of four banks of 40 Gabor filters, comprising eight different orientations and five scales.

{\bf Sample selection:} Positive samples were taken to be frames where the AU was present, and negative samples where it was not.
Since AU occurrence is relatively sparse among the entire video, we randomly downsampled to $\sim$20\% of the data with a 1:10 positive/negative ratio. Note that although better sampling strategies are possible (\eg, \cite{zhu2011dynamic,zeng2015confidence}), in the experiments we observed comparable performance using the DR techniques.

\begin{table}
	\caption{Running time and complexities of DR techniques}
	\label{tab:complexity}
	\vspace{-8pt}
	\centering
	\footnotesize
	\begin{tabular}{@{}r|cccc}
		\toprule
		Methods & PCA & KPCA & LLE & LPP \\
		\midrule
		Time & 0.565s & 7.048s & 4.012s & 1.922s \\
		Computational & $\mathcal{O}(d^3)$ & $\mathcal{O}(n^3)$ & $\mathcal{O}(pn^2)$ & $\mathcal{O}(pn^2)$ \\
		Memory & $\mathcal{O}(d^2)$ & $\mathcal{O}(n^2)$ & $\mathcal{O}(pn^2)$ & $\mathcal{O}(pn^2)$ \\
		\midrule\midrule
		Methods & LDA & KDA & \multicolumn{1}{@{}c@{}|}{LSDA} & \\
		\cmidrule(r){1-4}
		Time &  0.594s & 19.029s & \multicolumn{1}{@{}c@{}|}{7.018s}  & \multicolumn{1}{l@{}}{$p: \#$neighbors} \\
		Computational & $\mathcal{O}(dnt+t^3)$ & $\mathcal{O}(n^3)$ & \multicolumn{1}{@{}c@{}|}{$\mathcal{O}(pn^2)$} & \multicolumn{1}{l@{}}{$t: \min(d,n)$}\\
		Memory & $\mathcal{O}(d^2)$ & $\mathcal{O}(n^2)$ & \multicolumn{1}{@{}c@{}|}{$\mathcal{O}(pn^2)$} & \\
	\bottomrule
	\end{tabular}
	\vspace{-10pt}\\
\end{table}
\begin{table*}
	\caption{Results in RU-FACS (A and B) and Spectrum (C and D) for SIFT and Gabor features.}
	\label{tab:RUFACS:sift}
	\vspace{-10pt}
	\centering
	(A)
	\begin{tabular}{@{\ }c@{\ }||@{\ }c@{\ }|@{\ }c@{\ }@{\ }c@{\ }|@{\ }c@{\ }@{\ }c@{\ }|@{\ }c@{\ }|@{\ }c@{\ }@{\ }c@{\ }|@{\ }c@{\ }|@{\ }c@{\ }@{\ }c@{\ }|@{\ }c@{\ }@{\ }c@{\ }|@{\ }c@{\ }|@{\ }c@{\ }@{\ }c@{\ }}
		\toprule
		& \multicolumn{8}{@{\ }c|@{\ }}{Area Under the ROC Curve} & \multicolumn{8}{@{\ }c@{\ }}{F1 Score}\\ 
		\cmidrule{2-17}
		SIFT & --- & \multicolumn{2}{@{\ }c|@{\ }}{Linear} & \multicolumn{2}{@{\ }c|@{\ }}{Manifold} & \multicolumn{1}{@{\ }c|@{\ }}{Hybrid} & \multicolumn{2}{@{\ }c|@{\ }}{Supervised}
		& --- & \multicolumn{2}{@{\ }c|@{\ }}{Linear} & \multicolumn{2}{@{\ }c|@{\ }}{Manifold} & \multicolumn{1}{@{\ }c|@{\ }}{Hybrid} & \multicolumn{2}{@{\ }c@{\ }}{Supervised} \\
		\midrule
		\bf{AU} & {\bf{No-DR}} & {\bf{PCA}} & {\bf{KPCA}} & {\bf{LPP}} & {\bf{LLE}} & {\bf{LSDA}} & {\bf{LDA}} & {\bf{KDA}} 
		        & {\bf{No-DR}} & {\bf{PCA}} & {\bf{KPCA}} & {\bf{LPP}} & {\bf{LLE}} & {\bf{LSDA}} & {\bf{LDA}} & {\bf{KDA}}\\
		\midrule
		1  & .61 &.75 &.70 &.67 & .63&.71 &.64 &.60 &  .26& .44 & .42 & .38 & .21 & .35 & .34 & .20\\
		2  & .60 &.78 &.75 &.74 & .66&.66 &.69 &.58 &  .21& .50 & .35 & .49 & .31 & .46 & .42 & .17\\
		4  & .77 &.70 &.75 &.74 & .43&.76 &.81 &.70 &  .10& .12 & .12 & .15 & .10 & .13 & .13 & .20\\
		6  & .86 &.90 &.88 &.85 & .84&.79 &.85 &.89 &  .38& .45 & .48 & .29 & .43 & .35 & .38 & .54\\
		10 & .77 &.77 &.71 &.78 & .63&.74 &.78 &.71 &  .14& .11 & .12 & .10 & .17 & .13 & .15 & .15\\
		12 & .90 &.93 &.93 &.94 & .67&.88 &.91 &.91 &  .67& .73 & .68 & .73 & .37 & .64 & .68 & .68\\
		14 & .62 &.62 &.64 &.68 & .60&.72 &.61 &.70 &  .09& .15 & .15 & .16 & .10 & .13 & .12 & .12\\
		15 & .78 &.81 &.85 &.81 & .66&.89 &.87 &.84 &  .30& .33 & .22 & .32 & .20 & .37 & .38 & .28\\
		17 & .66 &.80 &.83 &.79 & .72&.75 &.80 &.86 &  .21& .34 & .34 & .34 & .28 & .34 & .36 & .39\\ \midrule
		Avg& .73 &.78 &.78 &.78 & .65&.77 &.77 &.75 &  .26& .35 & .32 & .35 & .24 & .32 & .33 & .30\\
		\bottomrule
	\end{tabular}
	\vspace{10pt}
	
	(B)
	\begin{tabular}{@{\ }c@{\ }||@{\ }c@{\ }|@{\ }c@{\ }@{\ }c@{\ }|@{\ }c@{\ }@{\ }c@{\ }|@{\ }c@{\ }|@{\ }c@{\ }@{\ }c@{\ }|@{\ }c@{\ }|@{\ }c@{\ }@{\ }c@{\ }|@{\ }c@{\ }@{\ }c@{\ }|@{\ }c@{\ }|@{\ }c@{\ }@{\ }c@{\ }}
		\toprule
		Gabor & --- & \multicolumn{2}{@{\ }c|@{\ }}{Linear} & \multicolumn{2}{@{\ }c|@{\ }}{Manifold} & \multicolumn{1}{@{\ }c|@{\ }}{Hybrid} & \multicolumn{2}{@{\ }c|@{\ }}{Supervised}
			& --- & \multicolumn{2}{@{\ }c|@{\ }}{Linear} & \multicolumn{2}{@{\ }c|@{\ }}{Manifold} & \multicolumn{1}{@{\ }c|@{\ }}{Hybrid} & \multicolumn{2}{@{\ }c@{\ }}{Supervised} \\
		\midrule
		\bf{AU} & {\bf{No-DR}} & {\bf{PCA}} & {\bf{KPCA}} & {\bf{LPP}} & {\bf{LLE}} & {\bf{LSDA}} & {\bf{LDA}} & {\bf{KDA}} 
		        & {\bf{No-DR}} & {\bf{PCA}} & {\bf{KPCA}} & {\bf{LPP}} & {\bf{LLE}} & {\bf{LSDA}} & {\bf{LDA}} & {\bf{KDA}}\\
		\midrule
		1	 & .46 &.75 &.73&.74 & .55&.56 &.64 & .59 &  .17& .43 & .31 & .36 & .18 & .18 & .34 & .22 \\
		2	 & .46 &.71 &.62&.56 & .57&.69 &.69 & .65 &  .26& .33 & .20 & .17 & .18 & .37 & .42 & .22 \\
		4	 & .43 &.59 &.64&.50 & .52&.51 &.51 & .58 &  .03& .04 & .05 & .03 & .03 & .03 & .03 & .03 \\
		6	 & .81 &.83 &.77&.87 & .66&.82 &.83 & .84 &  .32& .35 & .34 & .50 & .18 & .47 & .49 & .36 \\
		10 & .65 &.53 &.64&.61 & .51&.74 &.73 & .67 &  .07& .06 & .15 & .08 & .06 & .15 & .13 & .09 \\
		12 & .87 &.80 &.83&.88 & .75&.89 &.91 & .83 &  .63& .47 & .58 & .63 & .42 & .65 & .68 & .57 \\
		14 & .47 &.51 &.53&.54 & .51&.48 &.61 & .58 &  .07& .08 & .09 & .08 & .07 & .08 & .12 & .11 \\
		15 & .67 &.73 &.62&.72 & .54&.71 &.80 & .72 &  .08& .15 & .07 & .20 & .05 & .14 & .29 & .12 \\
		17 & .77 &.73 &.75&.82 & .61&.85 &.75 & .71 &  .22& .26 & .20 & .36 & .09 & .32 & .35 & .15 \\ \midrule
		Avg& .62 &.69 &.68&.69 & .58&.69 &.72 & .69 &  .21& .24 & .22 & .27 & .14 & .27 & .32 & .21 \\
		\bottomrule
	\end{tabular}
	\vspace{10pt}

	(C) 
	\begin{tabular}{@{\ }c@{\ }||@{\ }c@{\ }|@{\ }c@{\ }@{\ }c@{\ }|@{\ }c@{\ }@{\ }c@{\ }|@{\ }c@{\ }|@{\ }c@{\ }@{\ }c@{\ }|@{\ }c@{\ }|@{\ }c@{\ }@{\ }c@{\ }|@{\ }c@{\ }@{\ }c@{\ }|@{\ }c@{\ }|@{\ }c@{\ }@{\ }c@{\ }}
		\toprule
		SIFT & --- & \multicolumn{2}{@{\ }c|@{\ }}{Linear} & \multicolumn{2}{@{\ }c|@{\ }}{Manifold} & \multicolumn{1}{@{\ }c|@{\ }}{Hybrid} & \multicolumn{2}{@{\ }c|@{\ }}{Supervised}
			& --- & \multicolumn{2}{@{\ }c|@{\ }}{Linear} & \multicolumn{2}{@{\ }c|@{\ }}{Manifold} & \multicolumn{1}{@{\ }c|@{\ }}{Hybrid} & \multicolumn{2}{@{\ }c@{\ }}{Supervised} \\
		\midrule
		\bf{AU} & {\bf{No-DR}} & {\bf{PCA}} & {\bf{KPCA}} & {\bf{LPP}} & {\bf{LLE}} & {\bf{LSDA}} & {\bf{LDA}} & {\bf{KDA}} 
		        & {\bf{No-DR}} & {\bf{PCA}} & {\bf{KPCA}} & {\bf{LPP}} & {\bf{LLE}} & {\bf{LSDA}} & {\bf{LDA}} & {\bf{KDA}}\\
		\midrule
		1	 & .69 &.58 &.58&.51 & .38&.58 &.68 &.66 &  .28& .24 & .23 & .20 & .21 & .23 & .30 & .29\\
		2	 & .72 &.67 &.58&.62 & .50&.71 &.70 &.69 &  .32& .31 & .21 & .23 & .19 & .29 & .31 & .27\\
		4	 & .67 &.69 &.74&.65 & .60&.67 &.70 &.76 &  .35& .38 & .45 & .37 & .27 & .34 & .39 & .48 \\
		6	 & .70 &.68 &.66&.71 & .46&.74 &.70 &.81 &  .22& .21 & .19 & .22 & .14 & .24 & .22 & .34 \\
		10 & .68 &.54 &.51&.44 & .44&.51 &.64 &.66 &  .43& .39 & .36 & .36 & .35 & .36 & .43 & .44 \\
		12 & .80 &.79 &.80&.79 & .74&.57 &.73 &.75 &  .60& .55 & .57 & .56 & .52 & .37 & .48 & .57 \\
		14 & .82 &.83 &.85&.84 & .67&.77 &.83 &.74 &  .58& .62 & .62 & .64 & .46 & .63 & .65 & .54 \\
		15 & .59 &.64 &.56&.52 & .52&.67 &.66 &.55 &  .18& .23 & .10 & .10 & .08 & .16 & .15 & .08 \\
		17 & .52 &.50 &.50&.43 & .55&.50 &.49 &.58 &  .38& .38 & .38 & .38 & .39 & .37 & .38 & .39\\ \midrule
		Avg& .69 &.66 &.64&.61 & .54&.64 &.68 &.69 &  .37& .37 & .35 & .34 & .29 & .33 & .37 & .38 \\
		\bottomrule
	\end{tabular}
	\vspace{10pt}
	
	(D)
	\begin{tabular}{@{\ }c@{\ }||@{\ }c@{\ }|@{\ }c@{\ }@{\ }c@{\ }|@{\ }c@{\ }@{\ }c@{\ }|@{\ }c@{\ }|@{\ }c@{\ }@{\ }c@{\ }|@{\ }c@{\ }|@{\ }c@{\ }@{\ }c@{\ }|@{\ }c@{\ }@{\ }c@{\ }|@{\ }c@{\ }|@{\ }c@{\ }@{\ }c@{\ }}
		\toprule
		Gabor & --- & \multicolumn{2}{@{\ }c|@{\ }}{Linear} & \multicolumn{2}{@{\ }c|@{\ }}{Manifold} & \multicolumn{1}{@{\ }c|@{\ }}{Hybrid} & \multicolumn{2}{@{\ }c|@{\ }}{Supervised}
			& --- & \multicolumn{2}{@{\ }c|@{\ }}{Linear} & \multicolumn{2}{@{\ }c|@{\ }}{Manifold} & \multicolumn{1}{@{\ }c|@{\ }}{Hybrid} & \multicolumn{2}{@{\ }c@{\ }}{Supervised} \\
		\midrule
		\bf{AU} & {\bf{No-DR}} & {\bf{PCA}} & {\bf{KPCA}} & {\bf{LPP}} & {\bf{LLE}} & {\bf{LSDA}} & {\bf{LDA}} & {\bf{KDA}} 
		        & {\bf{No-DR}} & {\bf{PCA}} & {\bf{KPCA}} & {\bf{LPP}} & {\bf{LLE}} & {\bf{LSDA}} & {\bf{LDA}} & {\bf{KDA}}\\
		\midrule
		1	 & .73 &.53 &.52&.68 & .42&.61 &.67 & .52 &  .29& .21 & .22 & .30 & .21 & .25 & .30 & .22 \\
		2	 & .81 &.56 &.70&.63 & .50&.73 &.70 & .65 &  .57& .20 & .31 & .24 & .19 & .33 & .27 & .27 \\
		4	 & .82 &.81 &.81&.68 & .64&.73 &.79 & .52 &  .56& .51 & .45 & .34 & .33 & .42 & .49 & .25 \\
		6	 & .68 &.62 &.48&.60 & .47&.70 &.72 & .63 &  .20& .17 & .14 & .17 & .13 & .23 & .25 & .18 \\
		10 & .67 &.61 &.40&.53 & .48&.63 &.67 & .51 &  .42& .39 & .35 & .35 & .35 & .41 & .43 & .35 \\
		12 & .81 &.74 &.71&.74 & .58&.84 &.83 & .72 &  .57& .48 & .47 & .50 & .41 & .58 & .58 & .48 \\
		14 & .81 &.86 &.76&.83 & .49&.83 &.83 & .66 &  .58& .70 & .55 & .64 & .40 & .63 & .61 & .46 \\
		15 & .39 &.38 &.40&.41 & .58&.43 &.46 & .55 &  .07& .07 & .07 & .07 & .09 & .07 & .07 & .09 \\
		17 & .52 &.54 &.43&.54 & .49&.50 &.53 & .52 &  .38& .38 & .38 & .38 & .38 & .38 & .38 & .38 \\ \midrule
		Avg& .69 &.63 &.58&.63 & .53&.67 &.69 & .59 &  .40& .35 & .33 & .33 & .28 & .37 & .38 & .30 \\
		\bottomrule
	\end{tabular}
	\vspace{-1pt}
\end{table*}

{\bf Comparative methods:}
We implemented seven representative DR approaches:
\begin{noindlist}
	\item[$\bullet$] PCA, KPCA, LPP:  98\% energy was retained for each method, resulting in $k=247,251,248$, respectively.
	\item[$\bullet$] LLE: Again, 98\% energy was retained for each, and the number of neighbors was $p=12$, resulting in $k = 258$.
	\item[$\bullet$] LDA, KDA, LSDA: Because AU detection is posed as a binary problem, $\#\text{classes}=2$ and
	 $k=1$.  
\end{noindlist}

{\bf Classifiers:}
For each AU, separate linear SVM classifiers \cite{holarge} were used.
Parameter tuning was done using 5-fold subject-wise cross-validation on the training data.
For all methods, we chose the parameters that maximize the averaged cross validation F1 score.


{\bf Evaluation metrics:}
To evaluate performance, we used both Area Under the Curve (AUC) and F1, which is $\text{F1} = \frac{2\cdot\emph{Recall}\cdot\emph{Precision}}{\emph{Recall} + \emph{Precision}}$.  Both metrics are widely used in the literature, and they convey non-redundant information.  Area under the curve shows the relation between true and false positives.  F1 summarizes the trade-off between precision and recall.  


\setlength{\ww}{.4\linewidth}
\begin{figure*}[t]
	\centering
	\includegraphics[width=\ww,trim=160 260 180 250]{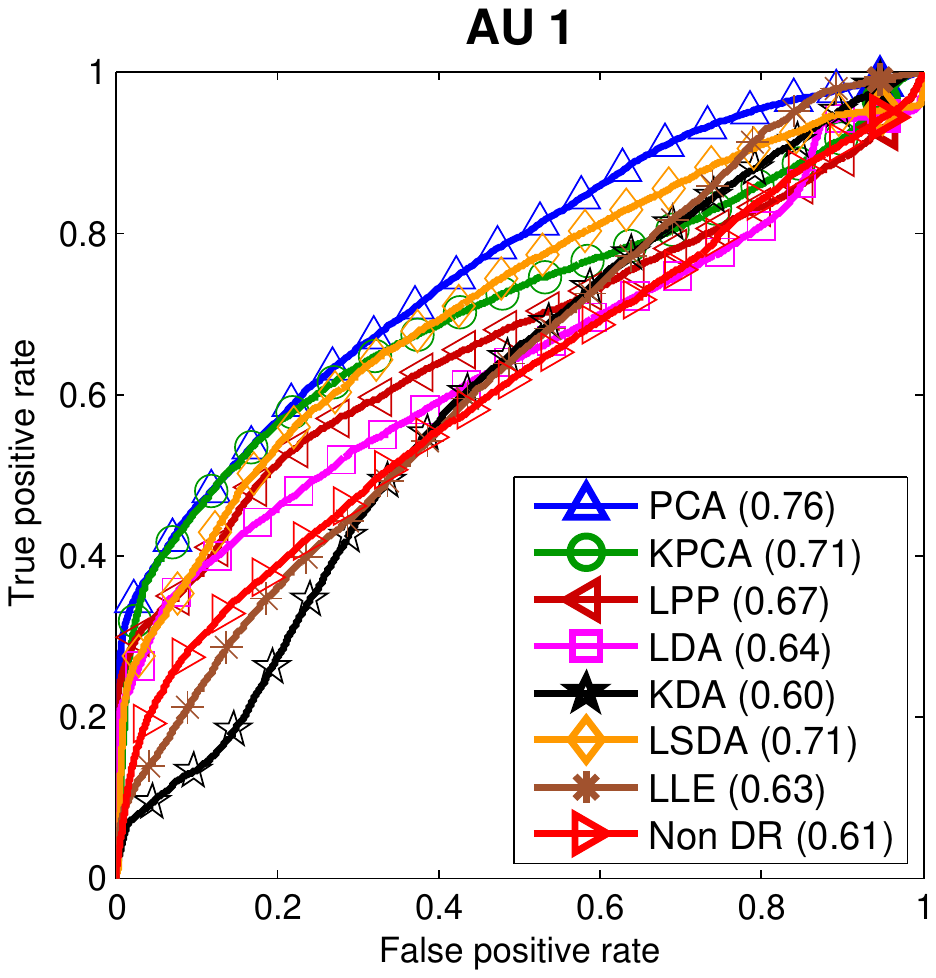}
	\includegraphics[width=\ww,trim=160 260 180 250]{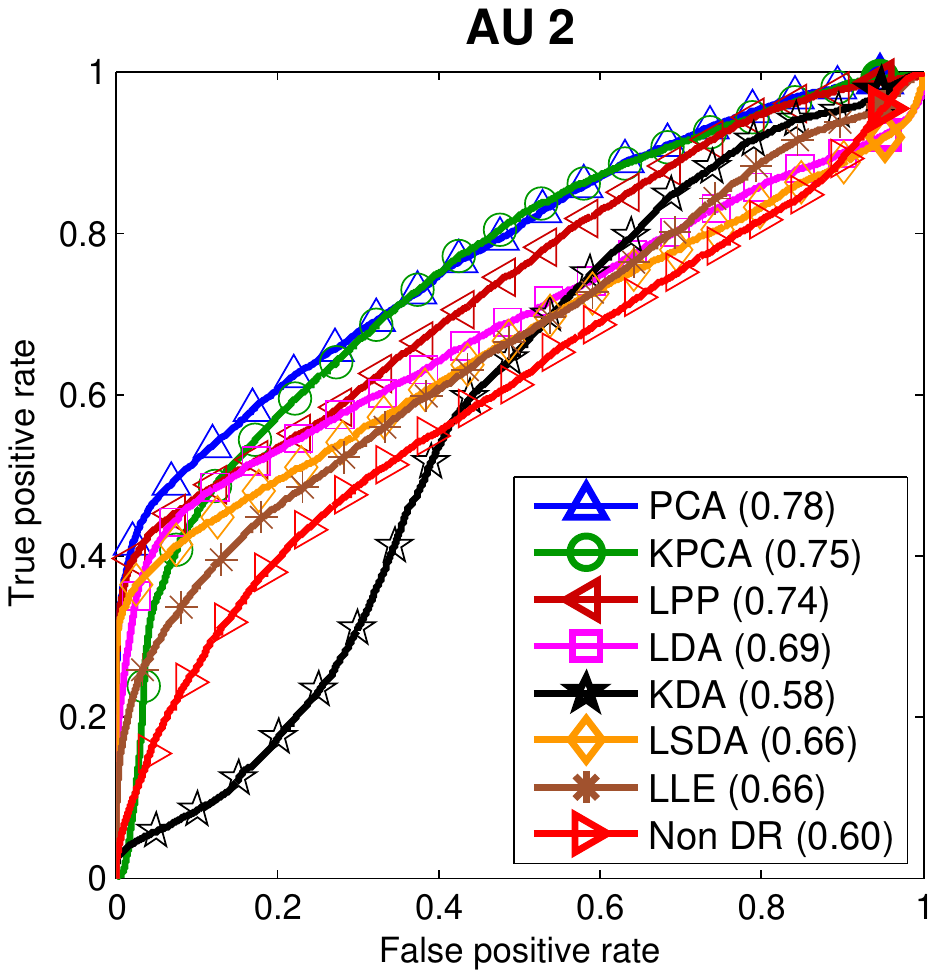}
	\includegraphics[width=\ww,trim=160 260 180 250]{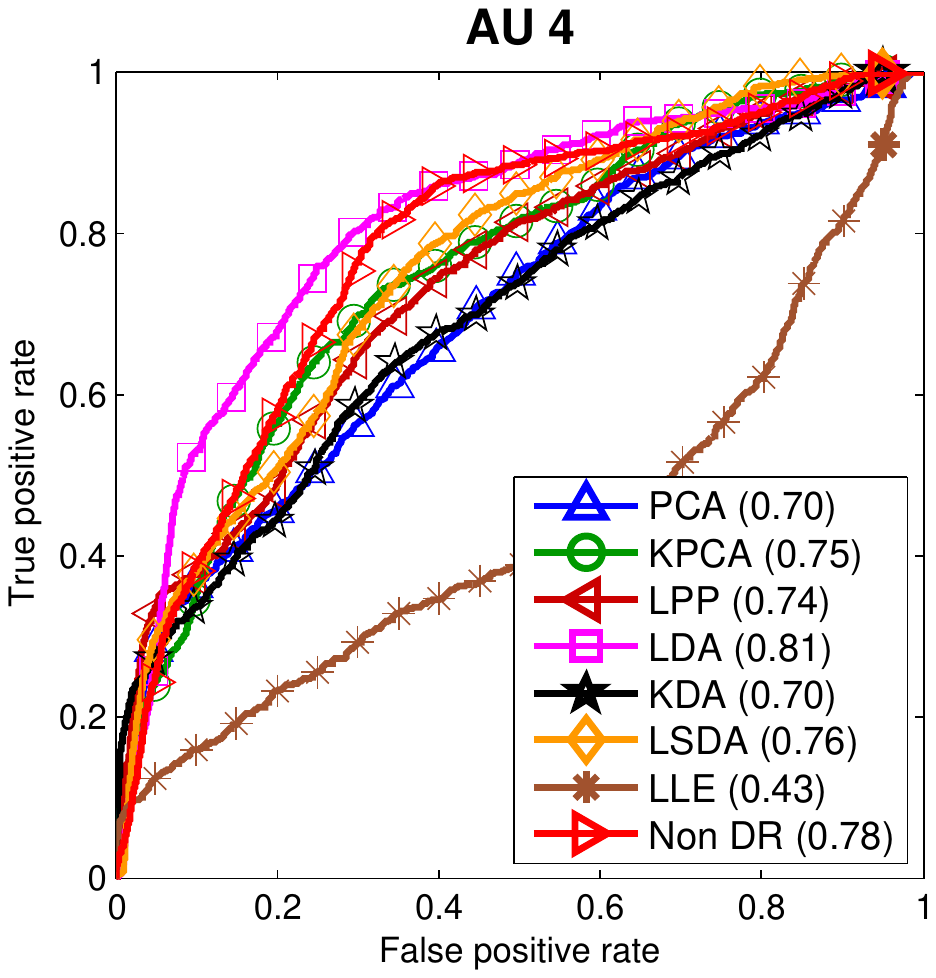}
	\includegraphics[width=\ww,trim=160 260 180 250]{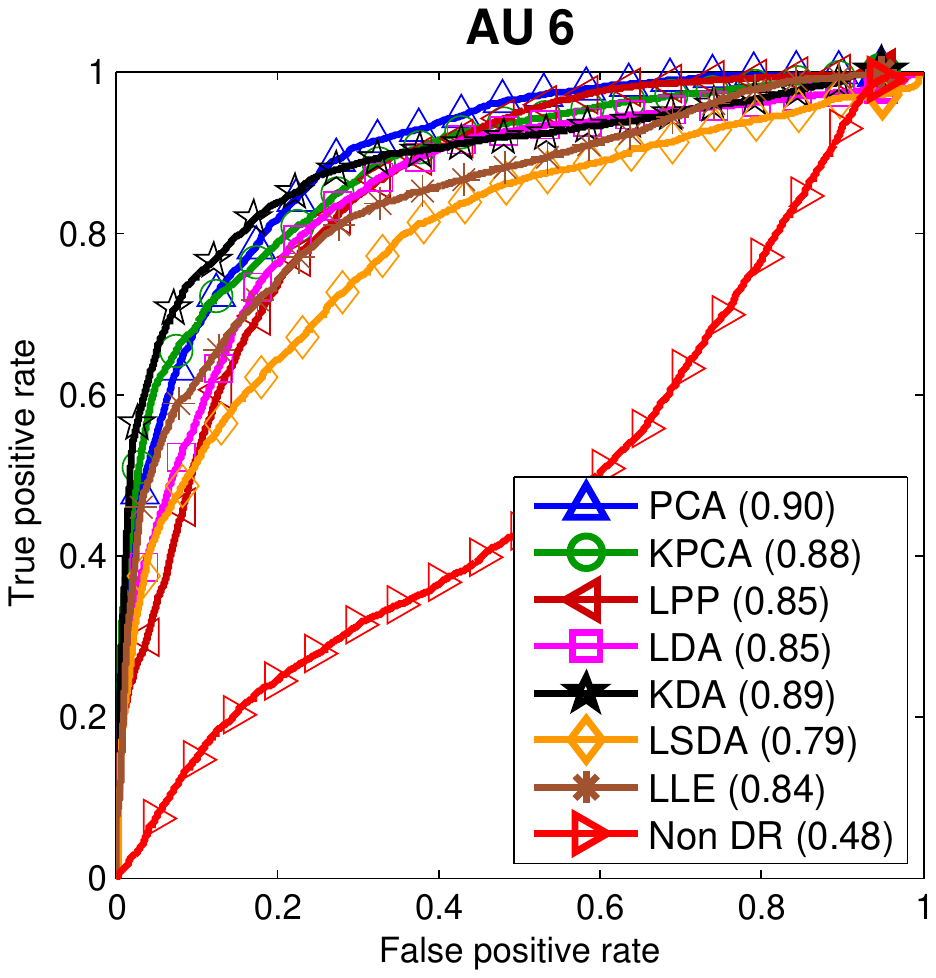}
	\includegraphics[width=\ww,trim=160 260 180 250]{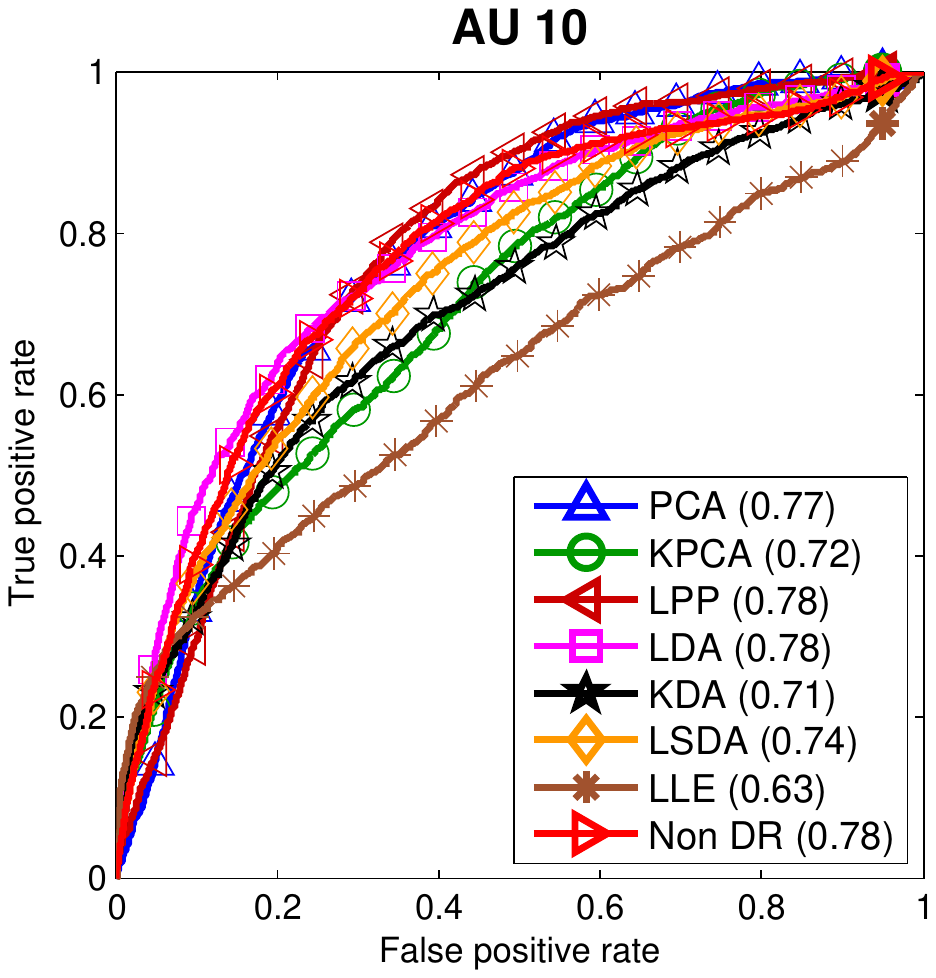}\\
	\includegraphics[width=\ww,trim=160 260 180 250]{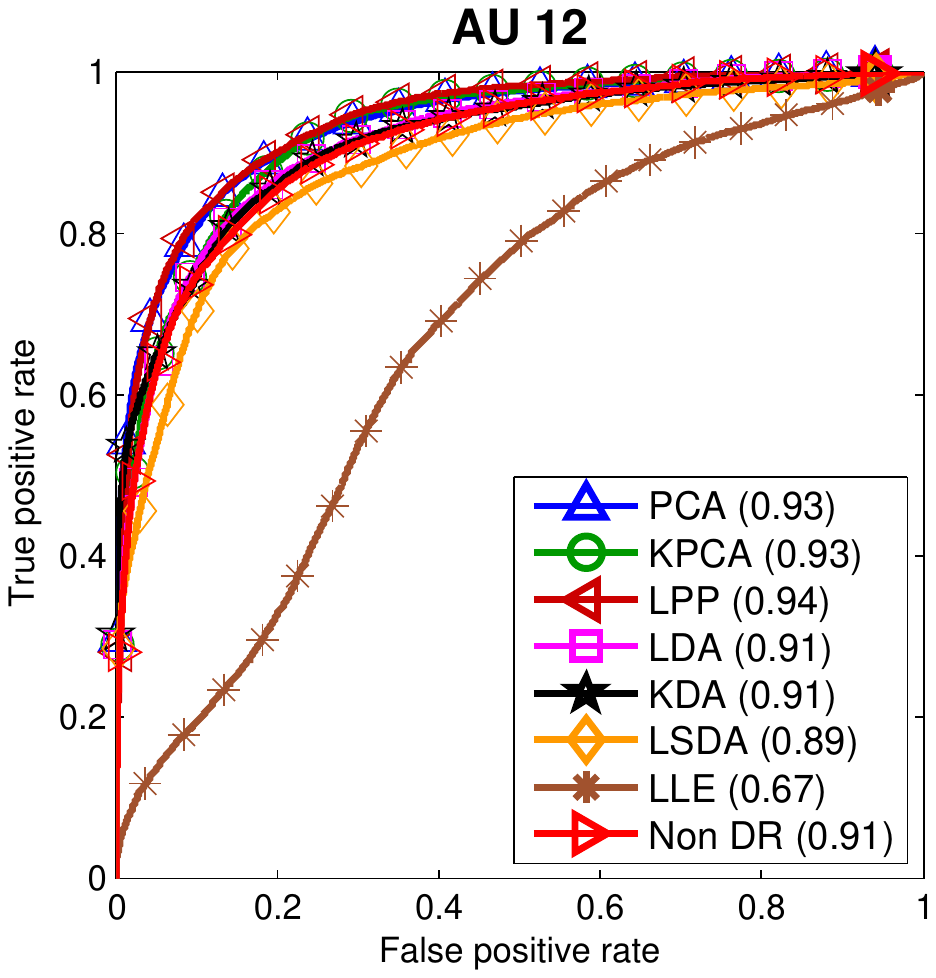}
	\includegraphics[width=\ww,trim=160 260 180 250]{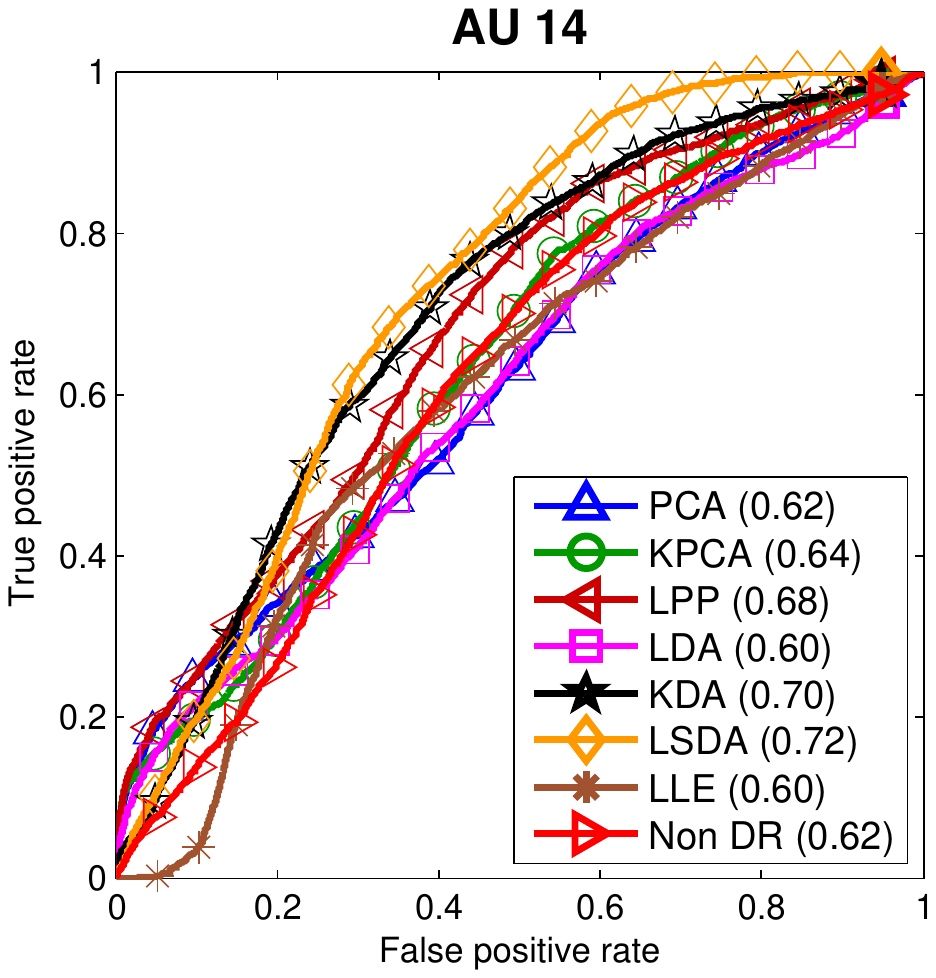}
	\includegraphics[width=\ww,trim=160 260 180 250]{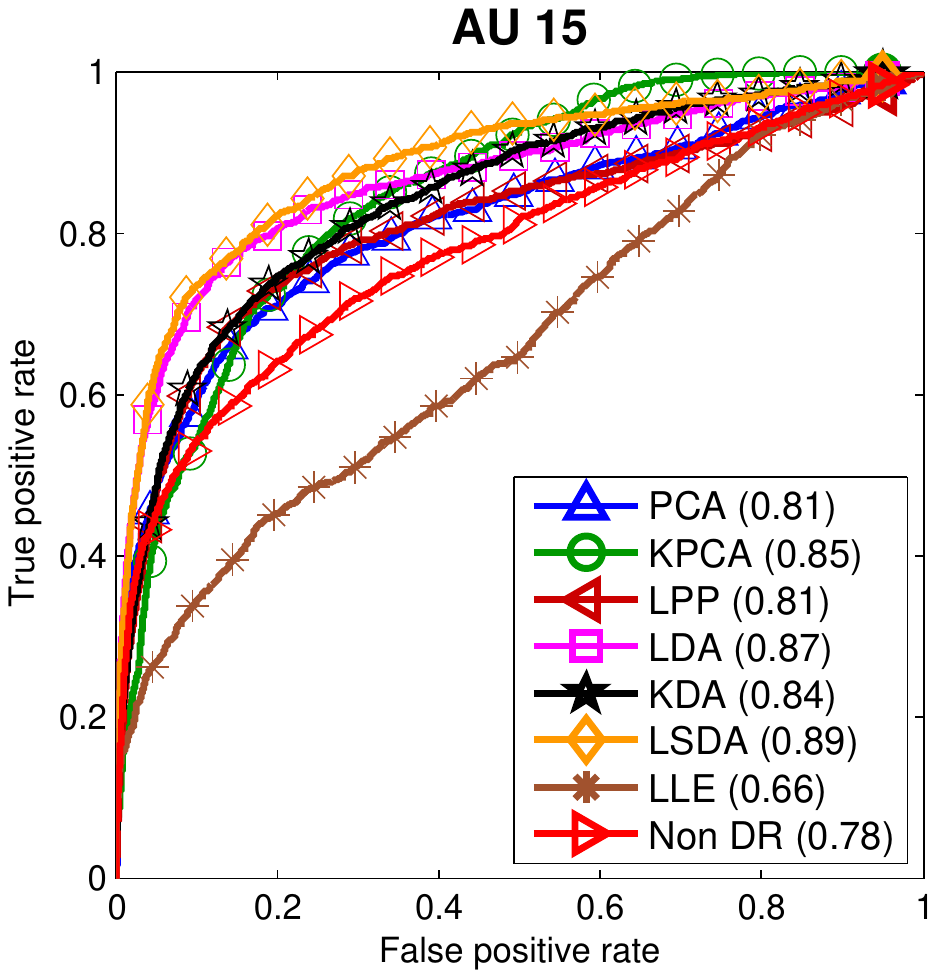}
	\includegraphics[width=\ww,trim=160 260 180 250]{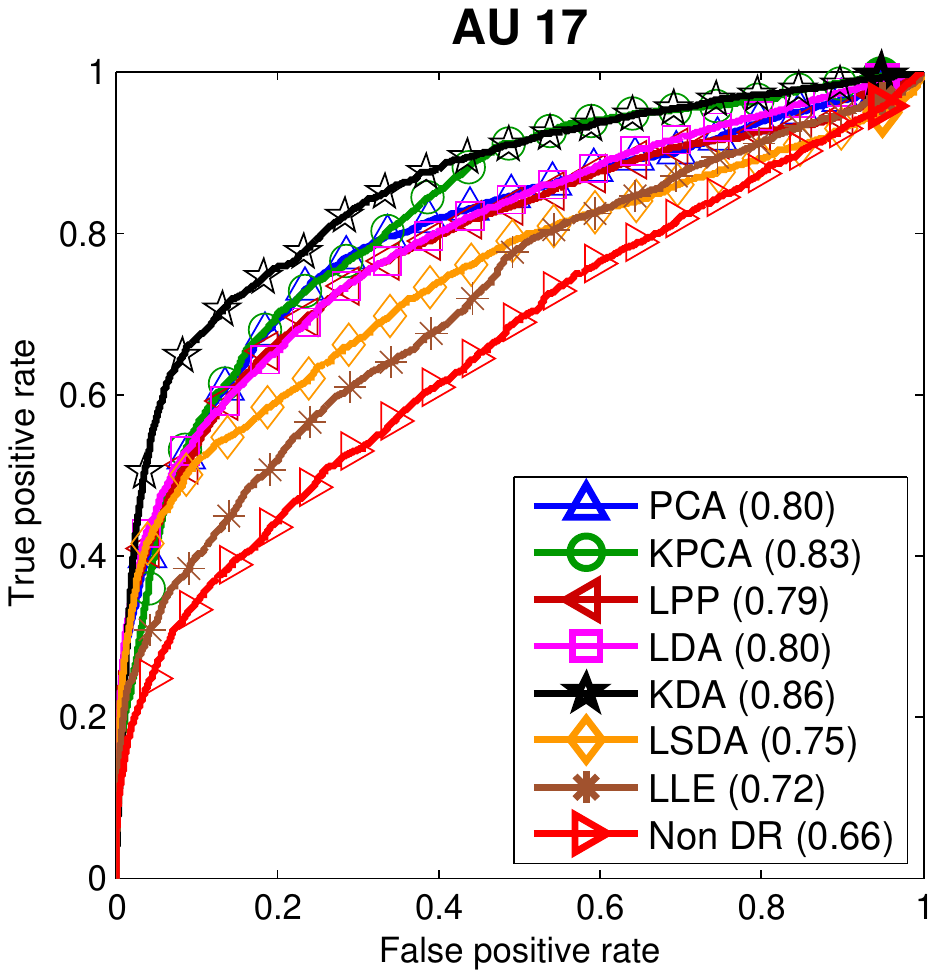}
	\vspace{-8pt}
	\caption{ROC curves on RU-FACS dataset using SIFT descriptors.}
	\label{fig:ROC:RUFACS:SIFT}
	\vspace{-8pt}
\end{figure*}

\begin{figure*}[t]
	\centering
	\includegraphics[width=\ww,trim=160 260 180 250]{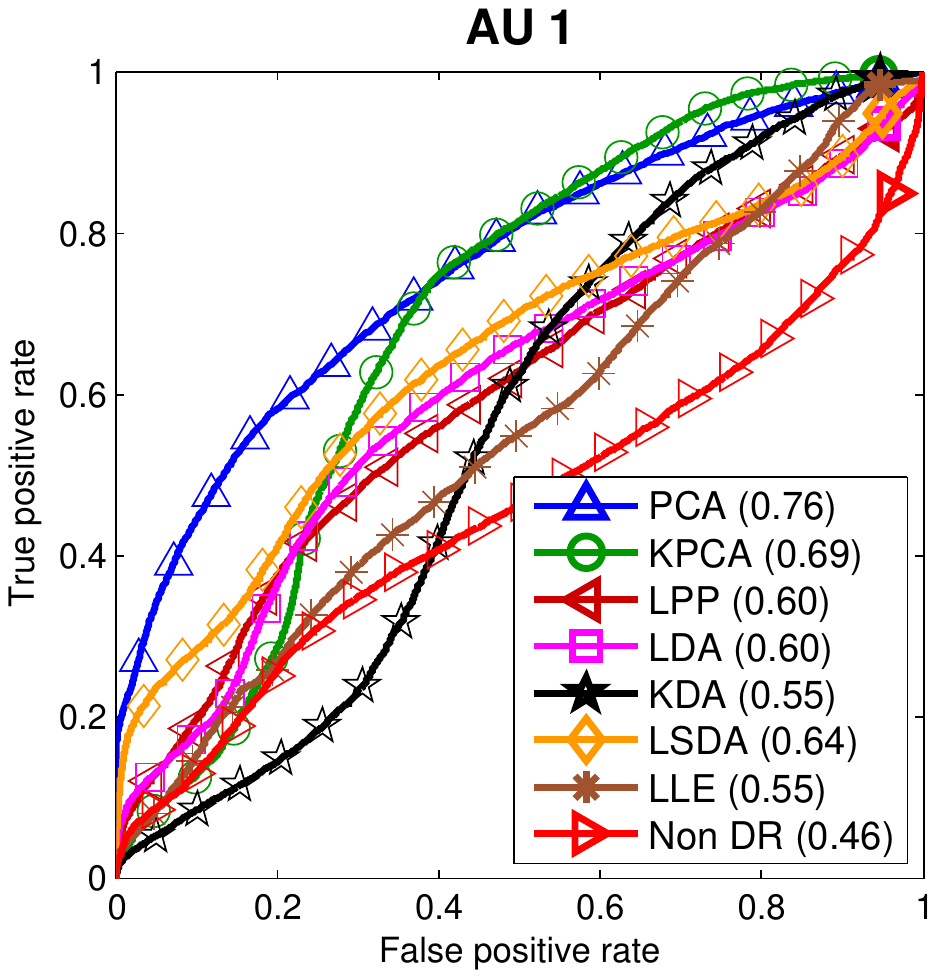}
	\includegraphics[width=\ww,trim=160 260 180 250]{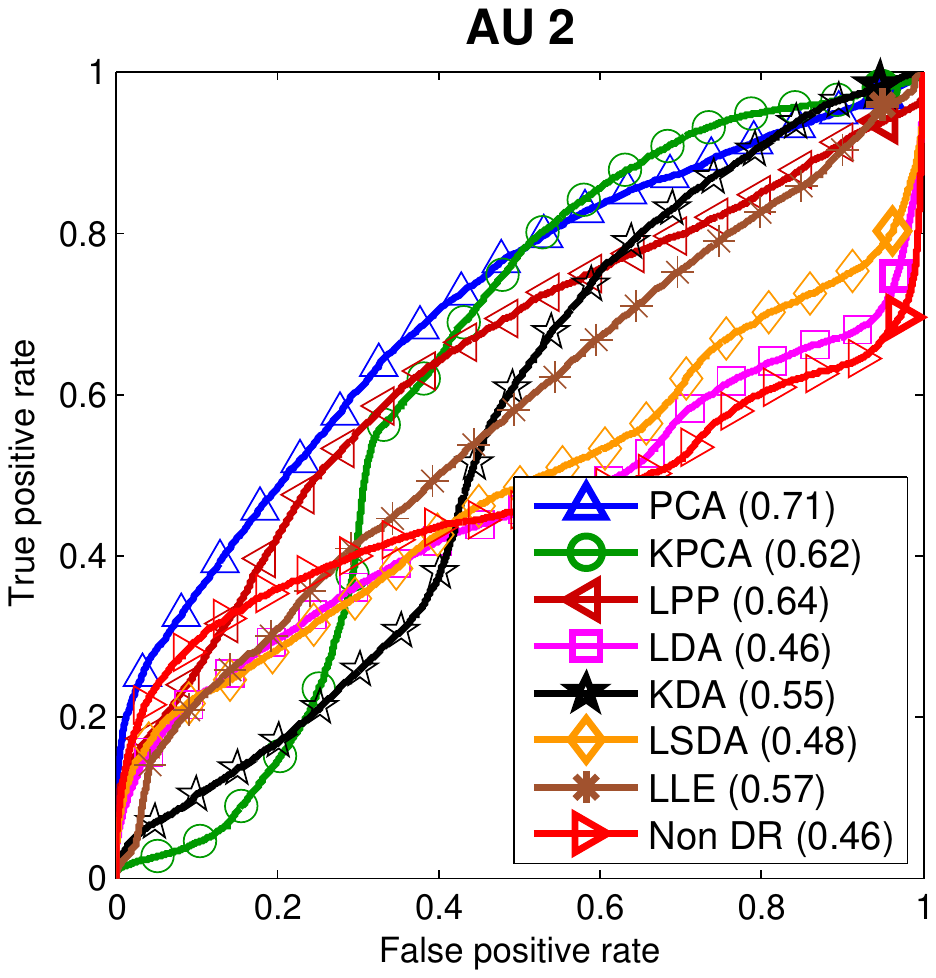}
	\includegraphics[width=\ww,trim=160 260 180 250]{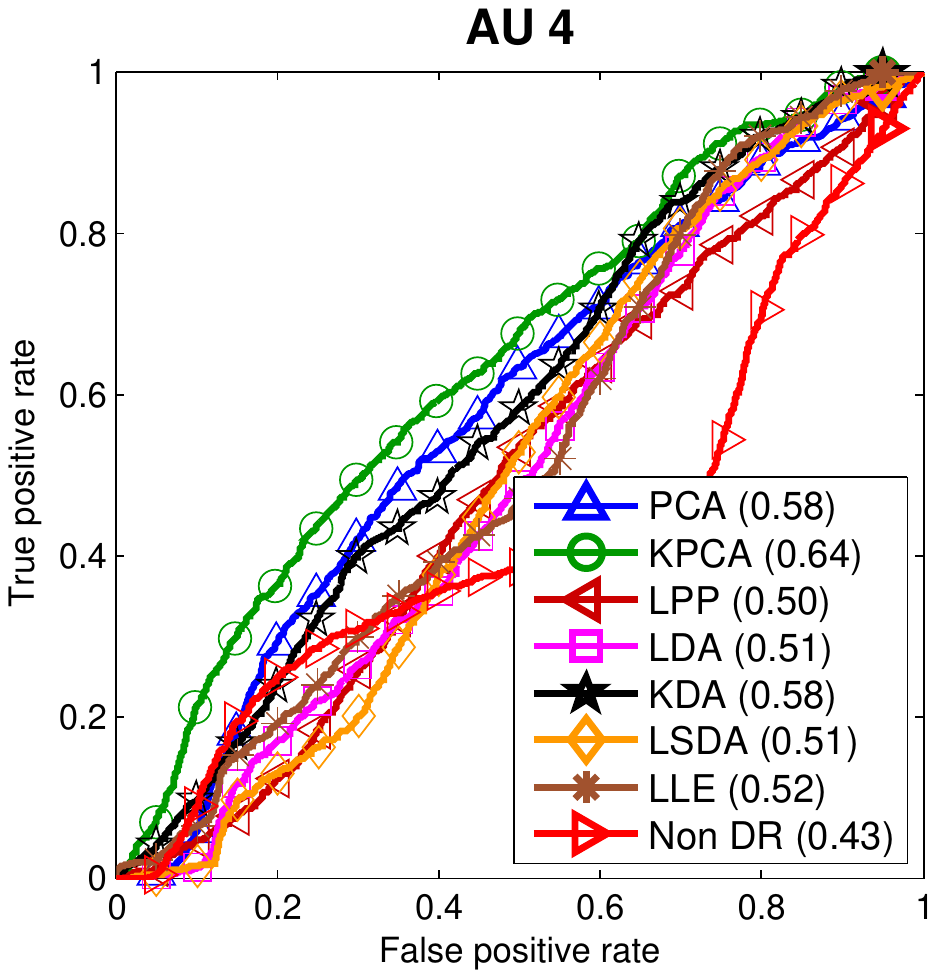}
	\includegraphics[width=\ww,trim=160 260 180 250]{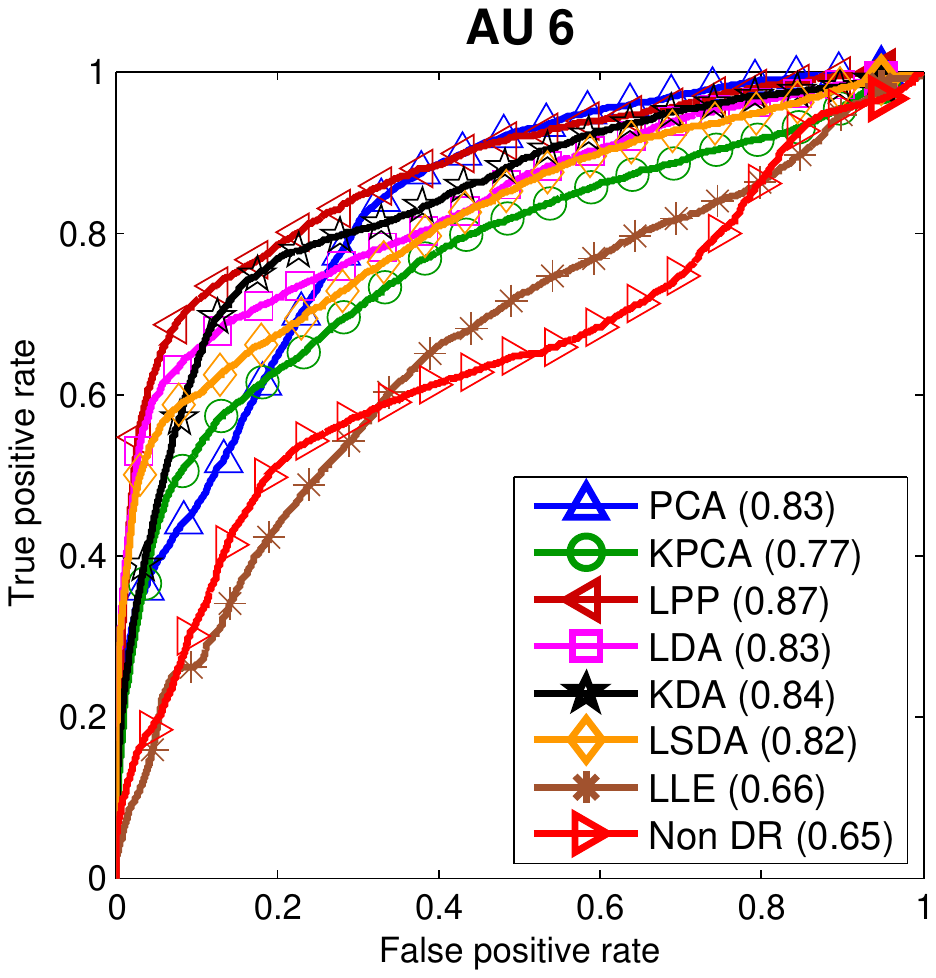}
	\includegraphics[width=\ww,trim=160 260 180 250]{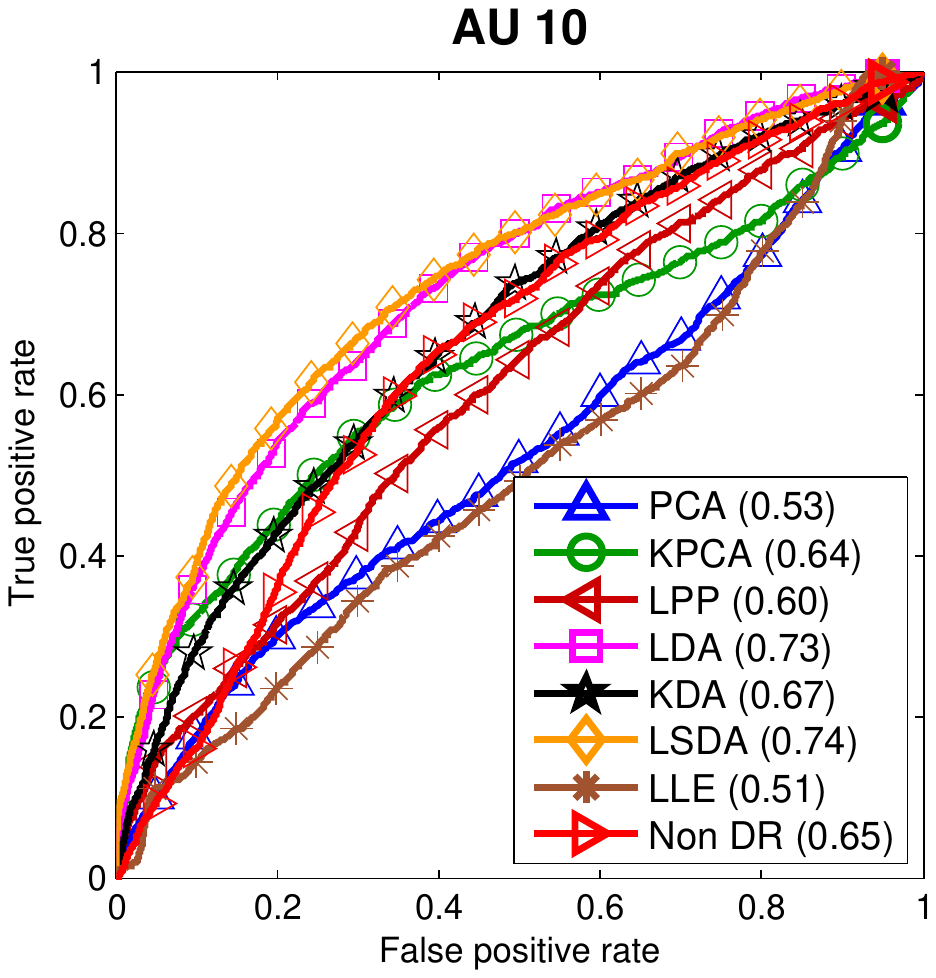}\\
	\includegraphics[width=\ww,trim=160 260 180 250]{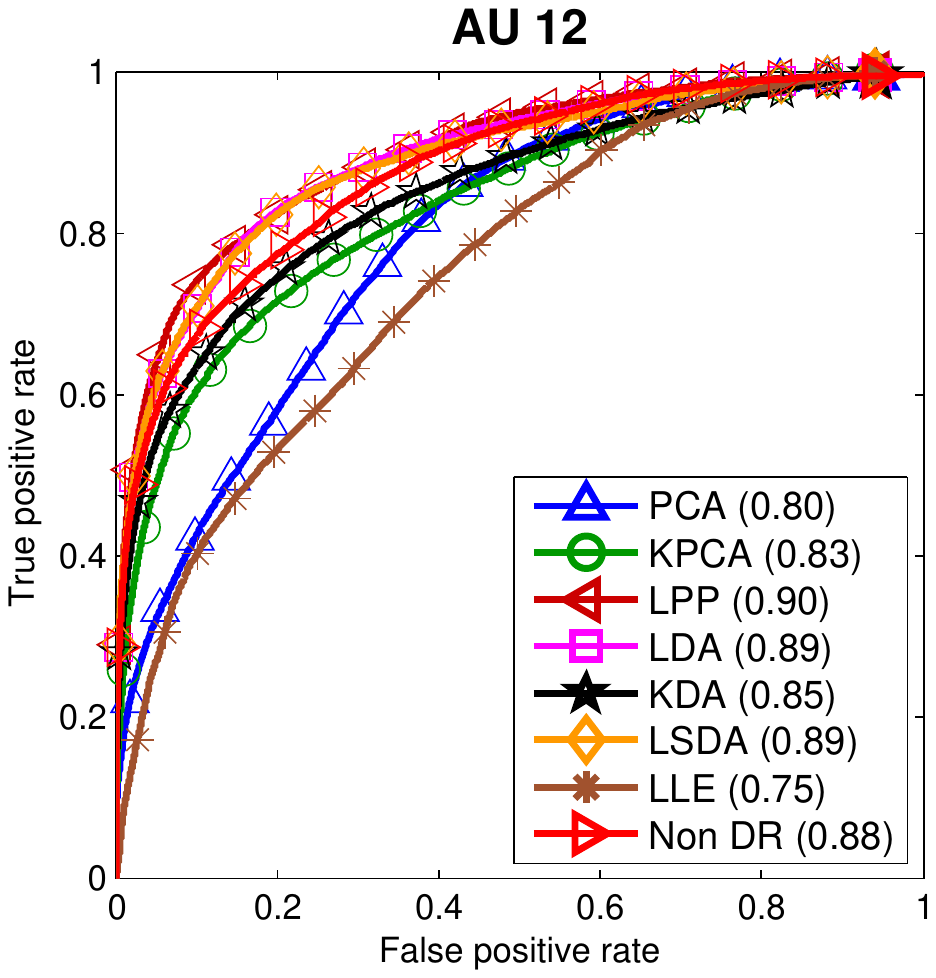}
	\includegraphics[width=\ww,trim=160 260 180 250]{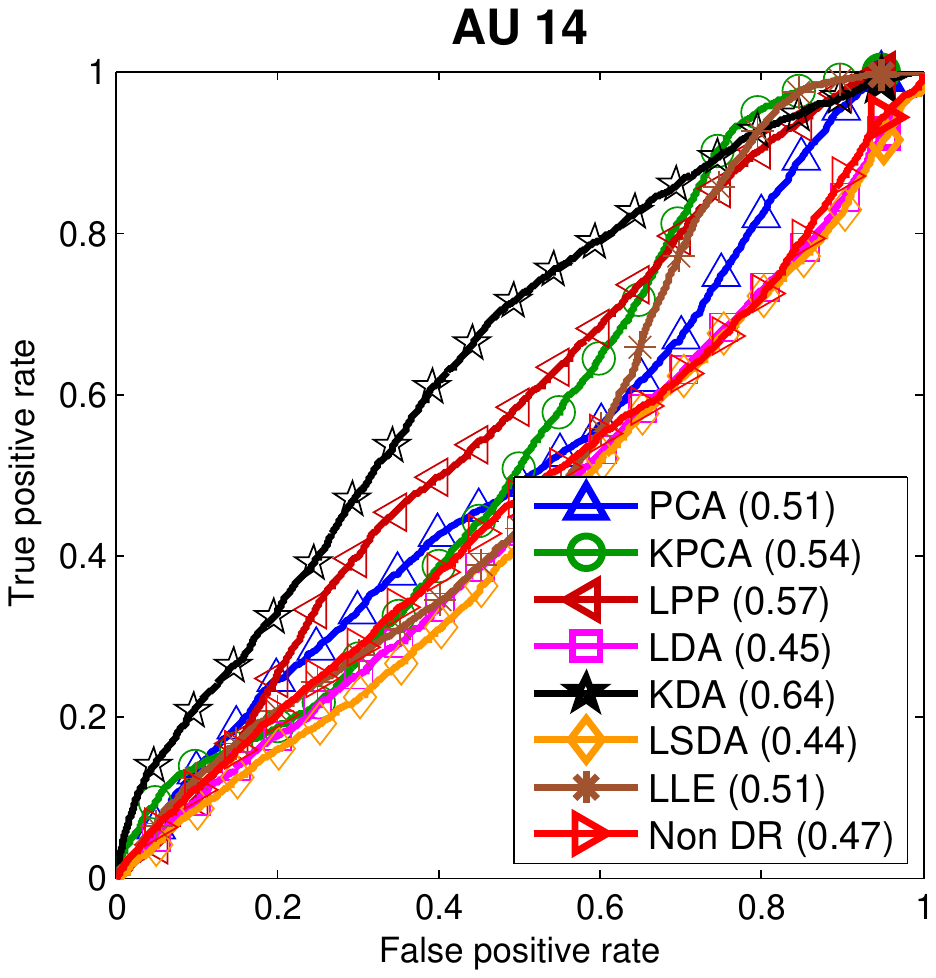}
	\includegraphics[width=\ww,trim=160 260 180 250]{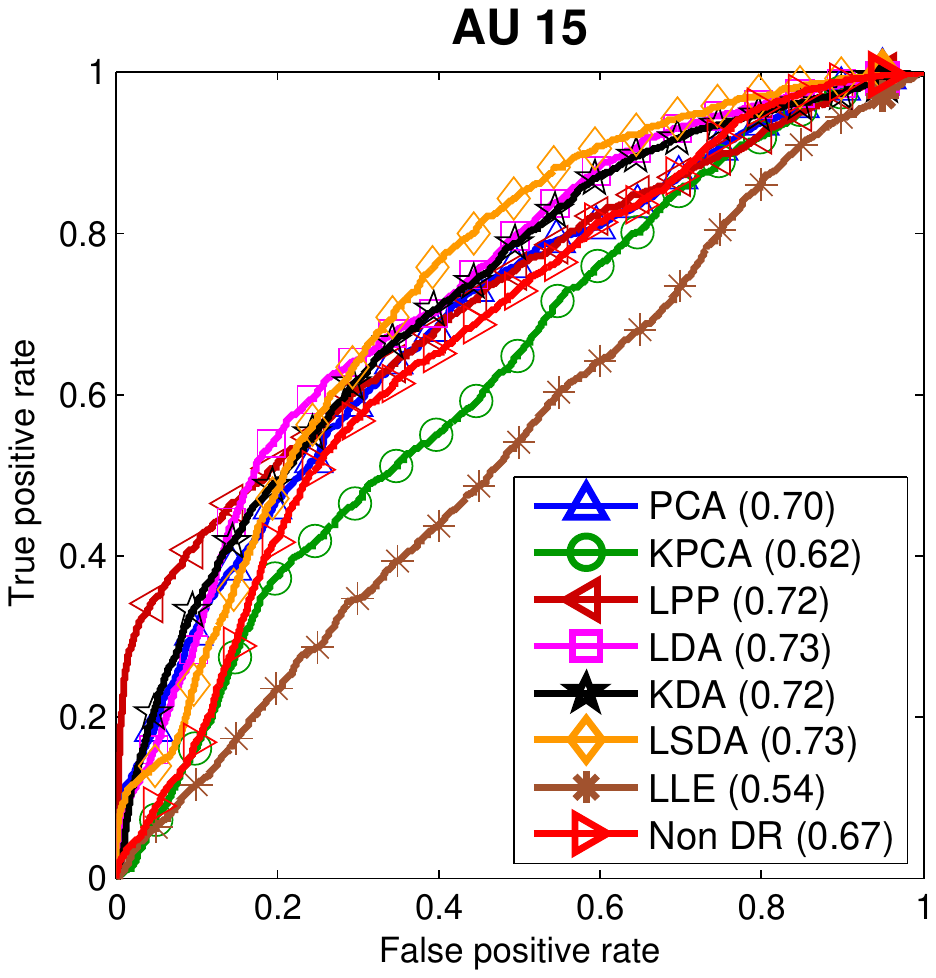}
	\includegraphics[width=\ww,trim=160 260 180 250]{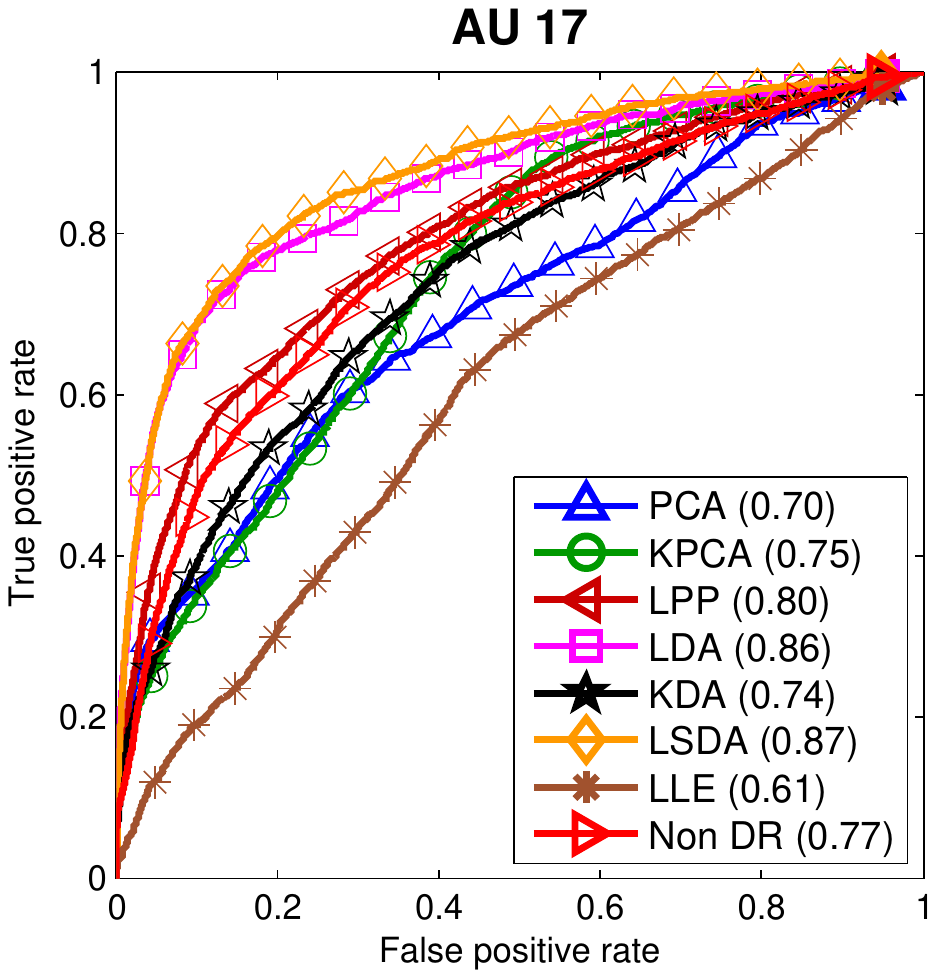}
	\vspace{-8pt}
	\caption{ROC curves on RU-FACS dataset using Gabor features.}
	\label{fig:ROC:RUFACS:Gabor}
	\vspace{-8pt}
\end{figure*}

\begin{figure*}[t]
	\centering
	\includegraphics[width=\ww,trim=160 260 180 250]{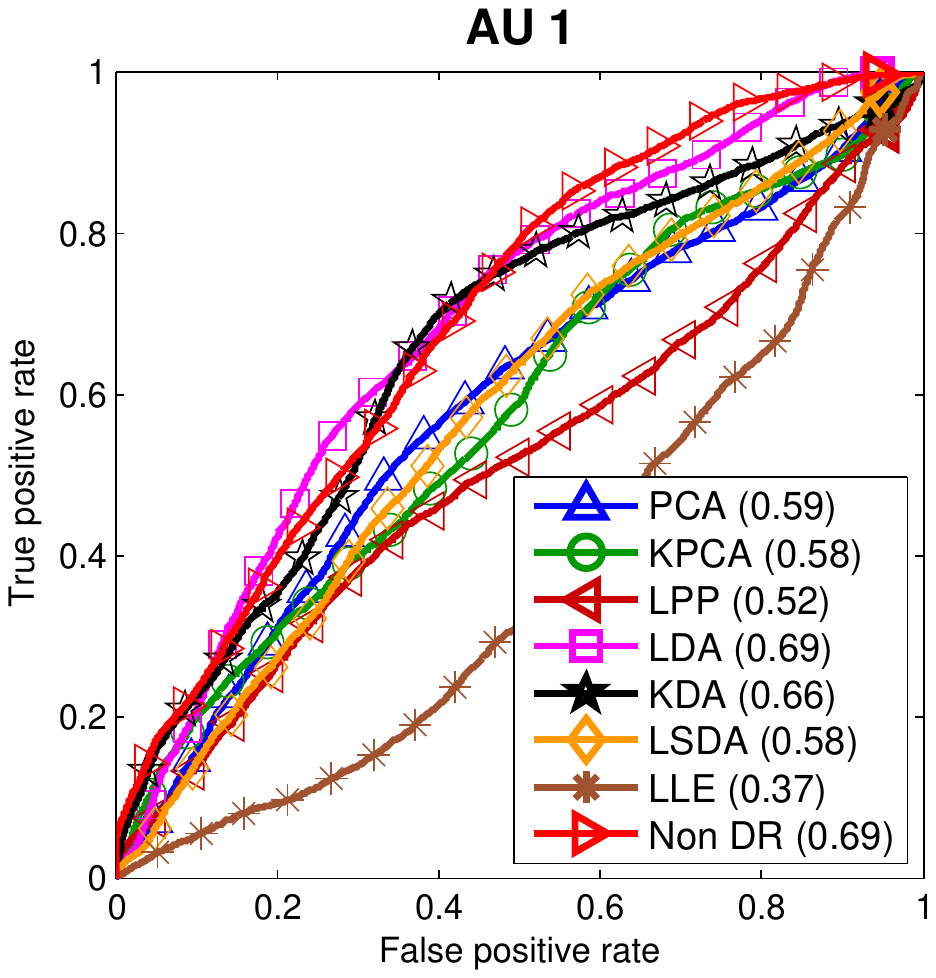}
	\includegraphics[width=\ww,trim=160 260 180 250]{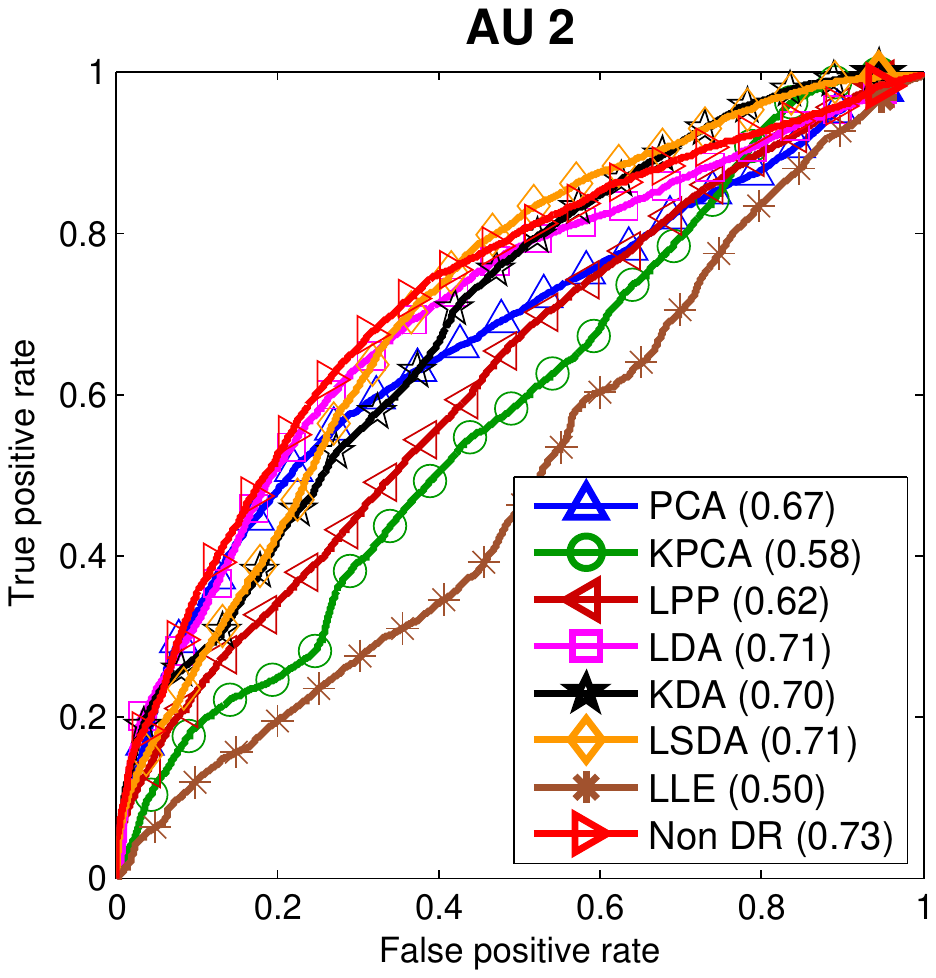}
	\includegraphics[width=\ww,trim=160 260 180 250]{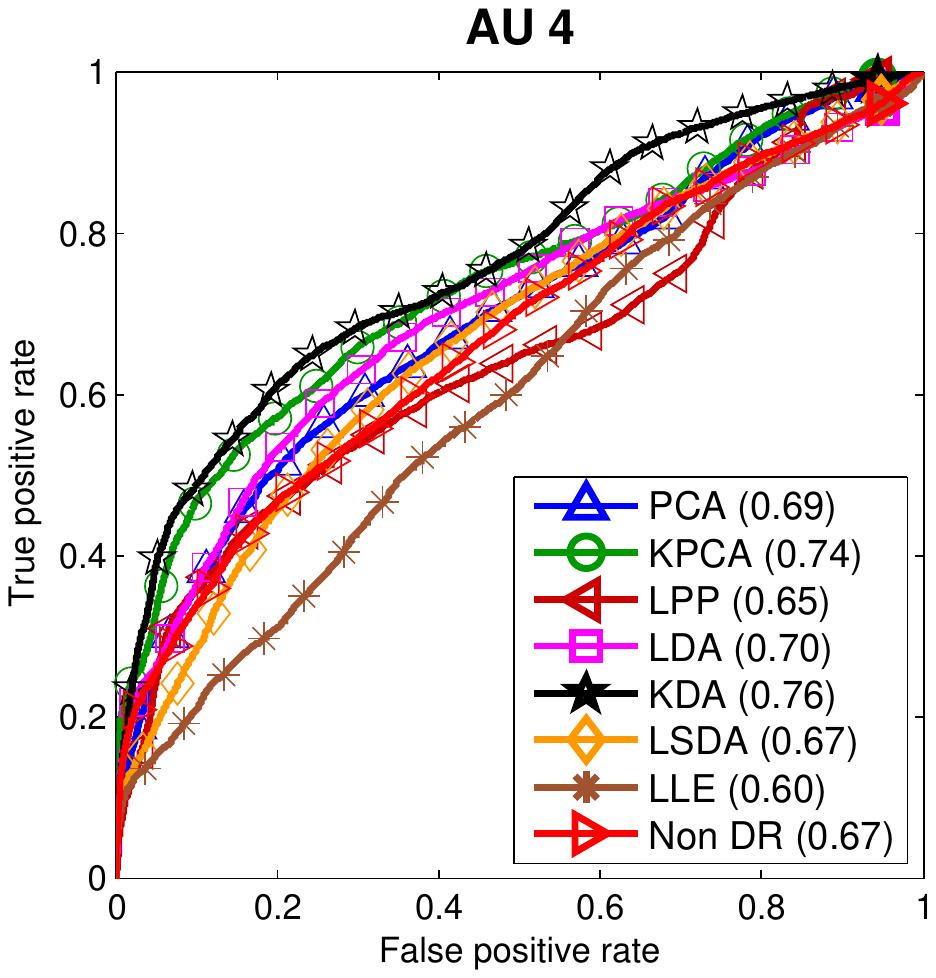}
	\includegraphics[width=\ww,trim=160 260 180 250]{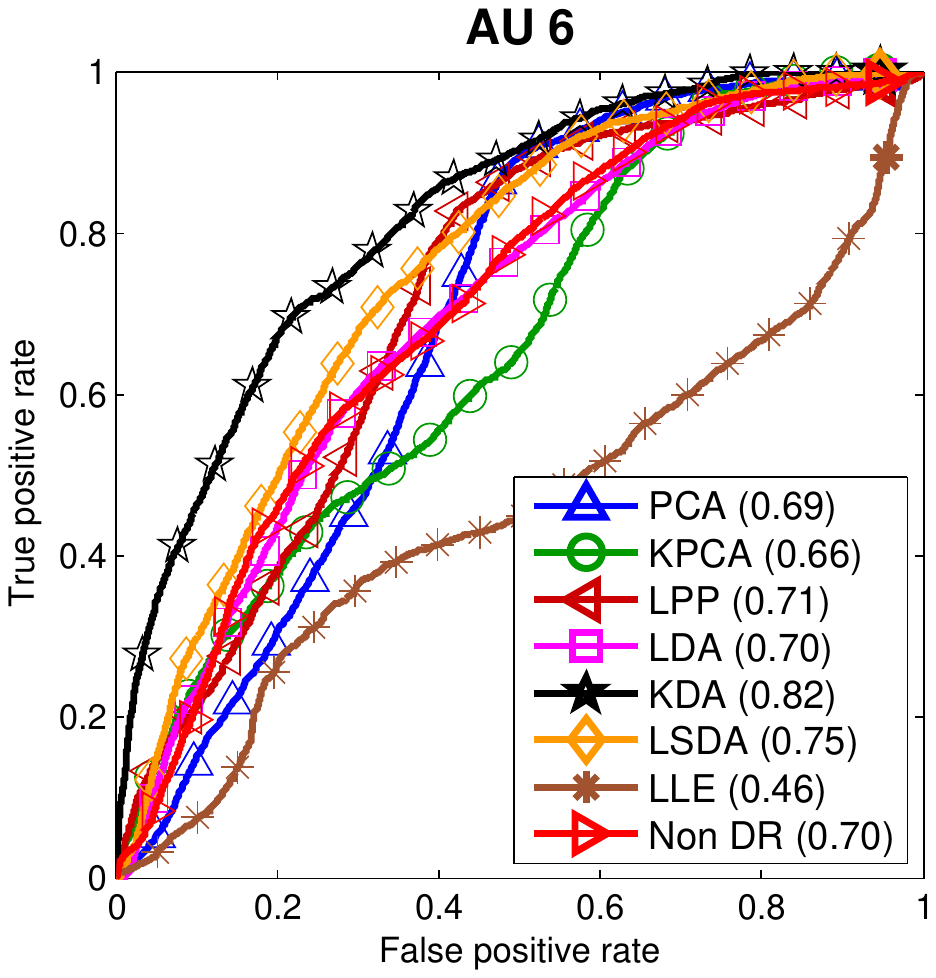}
	\includegraphics[width=\ww,trim=160 260 180 250]{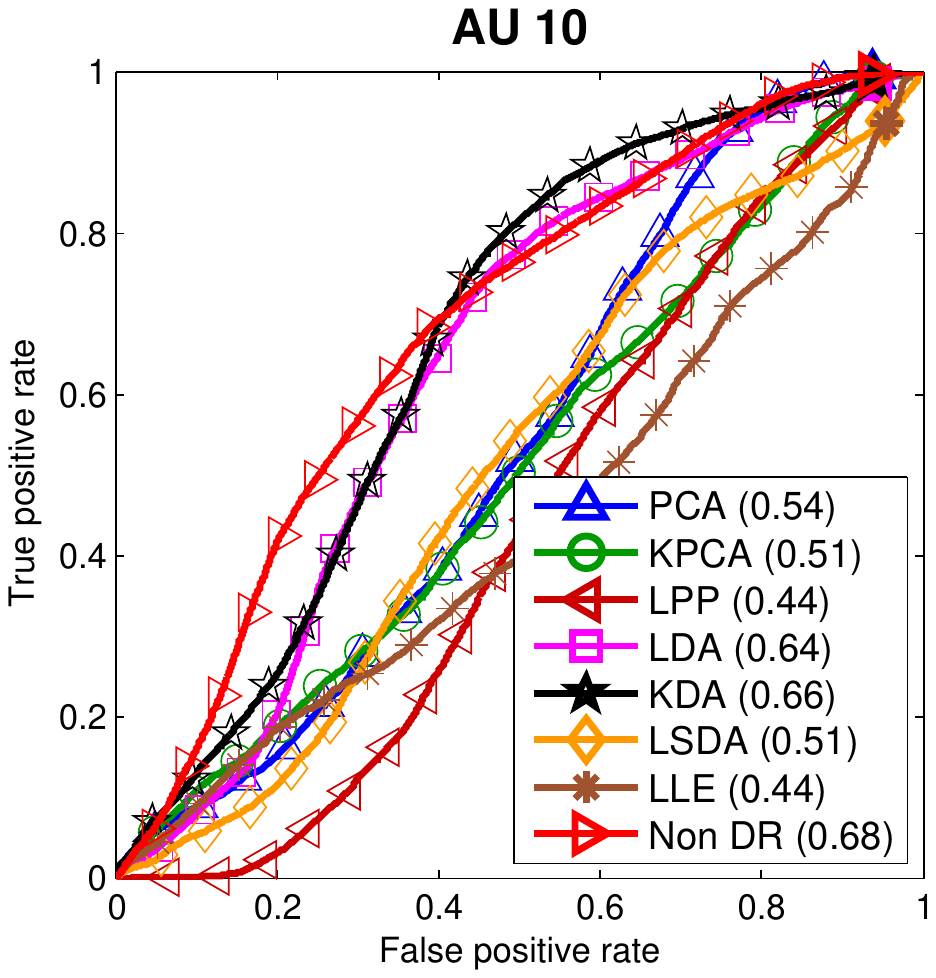}\\
	\includegraphics[width=\ww,trim=160 260 180 250]{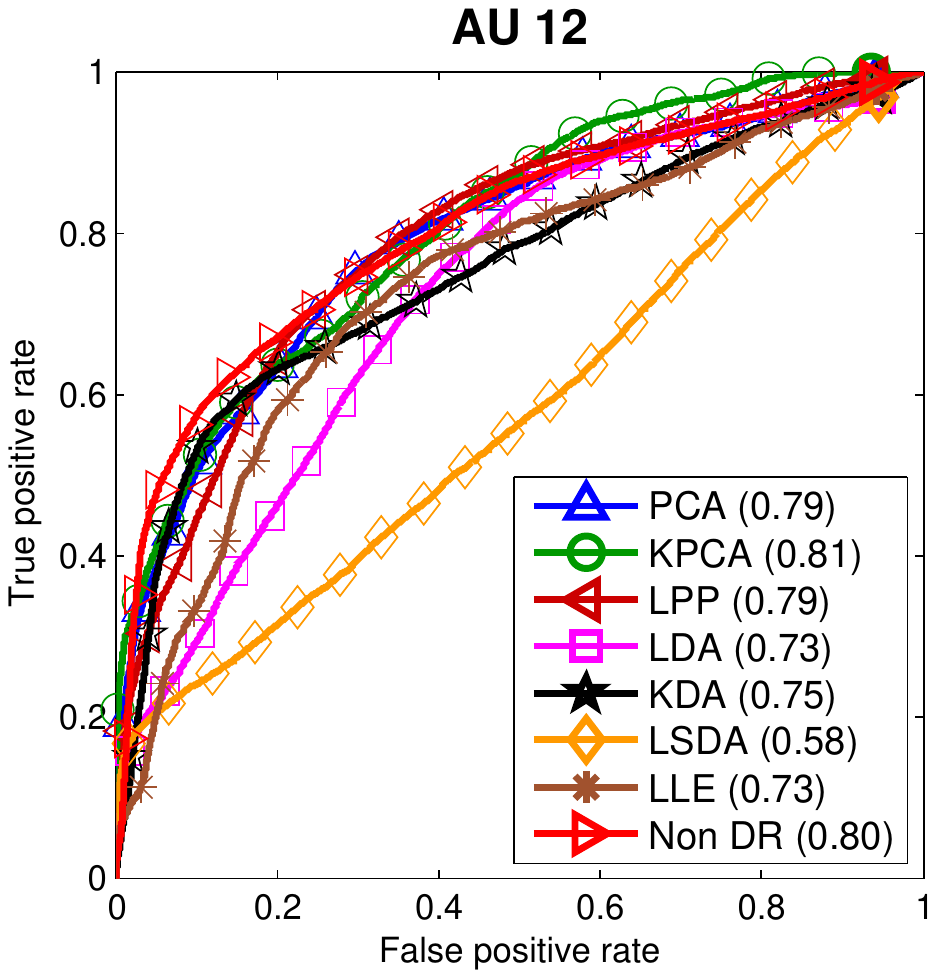}
	\includegraphics[width=\ww,trim=160 260 180 250]{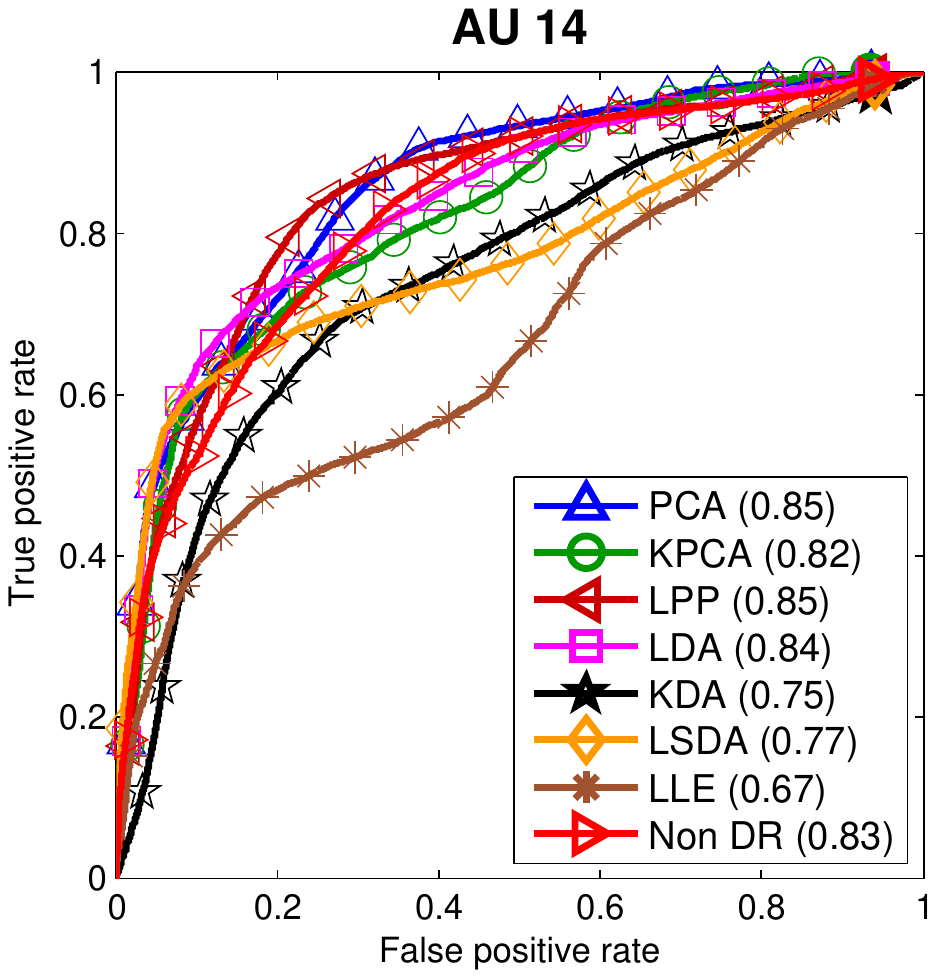}
	\includegraphics[width=\ww,trim=160 260 180 250]{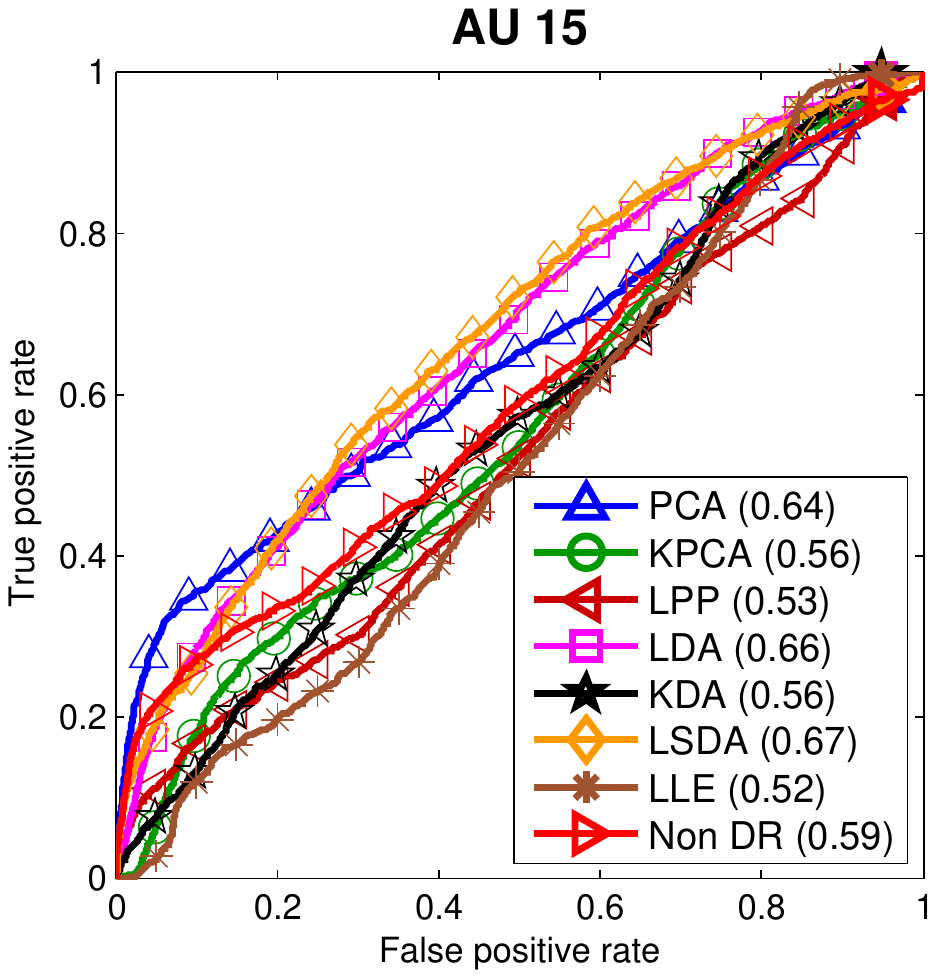}
	\includegraphics[width=\ww,trim=160 260 180 250]{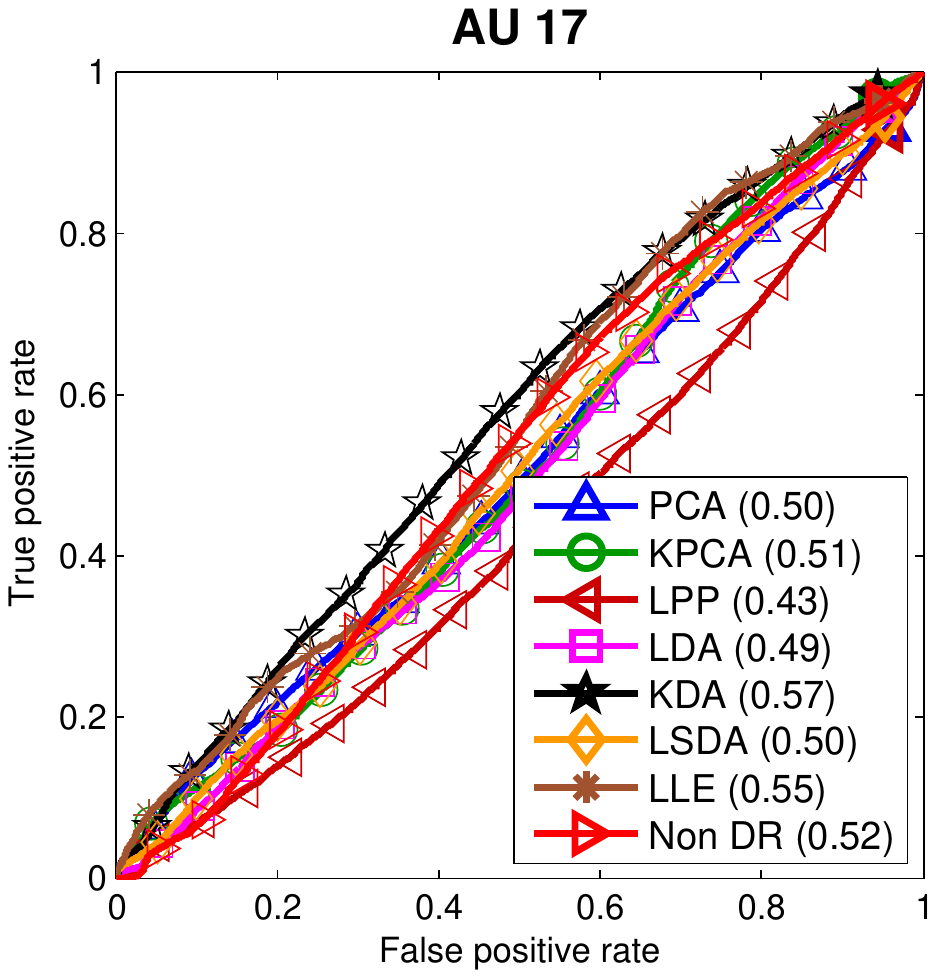}
	\vspace{-8pt}
	\caption{ROC curves on Spectrum dataset using SIFT features.}
	\label{fig:ROC:Spectrum:SIFT}
	\vspace{-8pt}
\end{figure*}


\subsection{Results and Discussions}

We first report findings for the relative efficiency of each DR technique.  We then report  findings with respect to features, appearance representations, and relative performance of each DR technique in both datasets. 
\subsubsection{Which approach is most efficient?}




Table \ref{tab:complexity} provides an overview of the computational demands and memory required for each DR techniques.
These results are relevant when considering the applicability and feasibility of each technique for use in a fielded system.
Though certain techniques may better approximate the original source space, they may be too resource intensive for some applications. 
To estimate run time, we ran a sub-experiment on 300 positive and 3,000 negative samples on an PC with CPU Intel i7 3.39GHz, Windows 7, and 8G RAM. 
The computational demands were primarily with respect to eigen-decomposition processes.
For kernel approaches, they involve an eigen-analysis of an $n\times n$ matrix and leads. 


\subsubsection{Lower-scale Gabor comparable with large-scale Gabor?}
\label{sec:disc:gabor}
Our system computed the Gabor responses only on a small subset of facial landmarks (we called it \emph{low-scale} Gabor), and \cite{bartlett2006automatic} computed Gabor on entire face images (\emph{large-scale} Gabor) and did not apply feature selection.
Note that, following \cite{zhu2011dynamic}, we used 60\%/40\% for selecting training/testing subjects, while \cite{bartlett2006automatic} used leave-one-subject-out strategy and hence retained more information in the training phase.
Observe in Table \ref{tab:RUFACS:sift}(b), with appropriate DR, our results on AUs 1 and 2 are more consistent (AUs 1 and 2 are similar upper-face movements) and achieves $\sim$5\% higher AUC than \cite{bartlett2006automatic}.
However, our system performed $\sim$15\% worse in AU 4 and $\sim$5\% worse on average.
A possible explanation is because AU 4 concentrates only on the inner portion of the eyebrow but we fixed the facial landmarks for every upper face AU.
We argue that, with suitable DR technique and landmark selection for different AUs, lower-scale Gabor can provide more reasonable results with greater efficiency.


\subsubsection{A better appearance representation?}
Following the settings on RU-FACS \cite{zhu2011dynamic}, we are able to replicate similar results, ensuring that our systems worked in a reasonable manner.
Observe from Tables \ref{tab:RUFACS:sift}, gradient-based SIFT consistently outperforms filter-based Gabor features across AUs and DR techniques, which has also been shown in \cite{zhu2011dynamic}.
This is because SIFT considers histograms of gradient orientations that potentially capture much FACS information such as naso-labial furrows and slope of eyebrows, and thus achieves the robustness to illumination changes and small errors in localization.
As a result, we converge to the conclusion that gradient-based SIFT leads to better AU detection than the filter-based Gabor features.
Moreover, we observed that the system achieved comparable performance with \cite{zhu2011dynamic} within $\sim$2\% AUC on average.
This indicates that DR techniques provide feature selection capability similar to bootstrapping for AU detection.

\subsubsection{DR affords advantages beyond efficiency?}
Despite the needs for DR (as mentioned in Sec.~\ref{sec:intro}), we explored further the benefits of DR approaches by including a no-DR control condition in the experiments.
Surprisingly, in Table \ref{tab:RUFACS:sift}, DR approaches generally improve not only the computational efficiency but also the performance.

\subsubsection{Which DR yield improved AU detection?}
In the study we compared four types of DR techniques including linear, manifold, supervised and hybrid methods.
There is not an overall winner across these types but we make the interesting observations:
(a) Kernel extensions do not always perform better.
Instead, in our experiments, PCA/LDA usually outperforms KPCA/KDA.
This can be explained by the \emph{curse of dimensionality} as the number of data grows exponentially with the dimensions in the feature space.
Lifting the data to a high-dimensional space requires construction of a proper kernel which leads to high computational costs.
Moreover, selection of parameters can be another issue, \eg, too small $\sigma$ for RBF kernel results in overfitting and large $\sigma$ introduces sensitivity to outliers.
(b) Manifold methods, considering neighborhood relationships, do not necessarily reflect the global AU structure.
The smoothness assumption of manifolds limits the modelling in the presence of noise and discontinuities in manifolds.
Data points face the difficulty to find good neighbors when the sampling space is sparse.
This is usually the case in AU detection since it is hard to represent the entire AU space with all possible face variations.
The poor performance of LLE in our experiments support this claims.
On the other hand, LPP is defined to cover the ambient space rather than just on the training samples (such as LLE), and thus can avoid the out-of-sample problem.
Observed from the experiments, the AUs with lots of samples can render the best performance by using LPP, such as AU 12 on RU-FACS, and AU 14 on Spectrum.
(c) Supervised methods generally help improve 3$\sim$6\% in AUC and up to 8\% in F1, as label information provides more discriminative properties for calculating the embedding.

\subsubsection{Results varied, so which DR for which AU?}
Observe from Tables \ref{tab:RUFACS:sift}, the pattern of results varied among AUs.
This motivates us to investigate the relationships between movements of AUs and the types of DR.
Take AUs 1/2 for example, the movements are to pull upwards the inner/outer portion of the eyebrows.
The muscle that triggers these movements can be interpreted as a linear function that produces vertical changes in the forehead.
As the movements are constrained, one can expect that linear DR methods should be adequate.
Our experiments show that in AUs 1 and 2, linear methods generally outperform manifold methods.
For complex AUs, especially lower face AUs involving asymmetric and spontaneous muscle movements, LPP works better than linear methods in general.

\section{Conclusion}
This paper presents a review and comparative study of techniques for DR techniques, 
which have not been devoted with much attention in AU detection.
Based on a unified LS-WKRRR framework, we have implemented a complete AU detection system incorporating seven representative DR techniques.
With the results on two popular appearance features SIFT and Gabor with two spontaneous datasets, we are able to answer several existing questions and gave our observations.
We also included a no-DR condition for further comparisons.
We have shown baseline results for PCA were comparable with previous research, and confirmed that gradient-based SIFT leads better AU detection than filter-based Gabor.
Manifold type of methods suffer from the assumption of good neighbors and are sensitive to noise and outliers.
Kernelized approaches are incapable of outperforming ordinary ones.


\bibliographystyle{ieee}
\bibliography{dr}

\end{document}